\documentclass[10pt,journal,compsoc]{IEEEtran}

\usepackage{times}
\usepackage{epsfig}
\usepackage{graphicx}
\usepackage{amsmath}
\usepackage{amssymb}
\usepackage{soul,color}
\usepackage[T1]{fontenc}
\usepackage{blindtext,nameref}
\usepackage{overpic}
\usepackage{booktabs}
\usepackage{multirow}
\usepackage{float}
\usepackage{stfloats}
\usepackage{marvosym}
\usepackage[breaklinks=true,bookmarks=false]{hyperref}
\usepackage[ruled,lined,linesnumbered]{algorithm2e}
\usepackage{amsfonts}

\newcommand{\tref}[1]{Table~\ref{#1}} %

\newcommand{\fref}[1]{Fig.~\ref{#1}}
\newcommand{\sref}[1]{Sec.~\ref{#1}}

\newcommand{\eg}{e.g.}

\newcommand{\hsf}[1]{\textcolor{black}{#1}}
\newcommand{\ysq}[1]{\textcolor{black}{#1}}
\newcommand{\lj}[1]{\textcolor{black}{#1}}

\newcommand{\frysq}[1]{\textcolor{black}{#1}}

\SetArgSty{textnormal}

\ifCLASSOPTIONcompsoc
  \usepackage[nocompress]{cite}
\else
  \usepackage{cite}
\fi

\ifCLASSINFOpdf
\else
\fi

\begin{document}
\title{Robust Point Cloud Segmentation with Noisy Annotations}

\author{Shuquan Ye, Dongdong Chen, Songfang Han, Jing Liao%
\IEEEcompsocitemizethanks{\IEEEcompsocthanksitem S. Ye is with the Department
        of Computer Science, City University of Hong Kong, Hong Kong SAR, China.\protect\\
        E-mail: shuquanye2-c@my.cityu.edu.hk
        \IEEEcompsocthanksitem D. Chen is with Microsoft Cloud+AI, Redmond, Washington, USA.\protect\\
        E-mail: cddlyf@gmail.com
        \IEEEcompsocthanksitem S. Han is with University of California, San Diego, USA.\protect\\
        E-mail: s5han@eng.ucsd.edu
        \IEEEcompsocthanksitem J. Liao is with the Department of Computer Science, City University of Hong Kong, Hong Kong SAR, China.\protect\\
        E-mail: jingliao@cityu.edu.hk
        \IEEEcompsocthanksitem J. Liao is the corresponding author.}
        }

\markboth{Journal of \LaTeX\ Class Files,~Vol.~14, No.~8, August~2015}%
{Shell \MakeLowercase{\textit{et al.}}: Bare Advanced Demo of IEEEtran.cls for IEEE Computer Society Journals}

\IEEEtitleabstractindextext{%
\begin{abstract}
  Point cloud segmentation is a fundamental task in 3D. Despite recent progress on point cloud segmentation with the power of deep networks, current learning methods based on the clean label assumptions may fail with noisy labels. 
  \ysq{Yet, class labels are often mislabeled at both instance-level and boundary-level in real-world datasets.}
  In this work, we take the lead in solving 
  \ysq{the instance-level label noise} by proposing a Point Noise-Adaptive Learning (PNAL) framework. Compared to noise-robust methods on image tasks, our framework is noise-rate blind, to cope with the spatially variant noise rate
  specific to point clouds. Specifically, we propose a point-wise confidence selection to obtain reliable labels from the historical predictions of each point. A cluster-wise label correction is proposed with a voting strategy to generate the best possible label by considering the neighbor correlations. \ysq{To handle boundary-level label noise, we also propose a variant ``PNAL-boundary " with a progressive boundary label cleaning strategy.} Extensive experiments demonstrate its effectiveness on both synthetic and real-world noisy datasets. 
  Even with $60\%$ symmetric noise \ysq{and high-level boundary noise}, our framework significantly outperforms its baselines, and is comparable to the upper bound trained on completely clean data. 
  Moreover, we cleaned the 
  popular real-world dataset ScanNetV2 for rigorous experiment. Our code and data \ysq{is available at https://github.com/pleaseconnectwifi/PNAL}.
\end{abstract}

\begin{IEEEkeywords}
Point Cloud, Scene Segmentation, Machine Learning, Noisy Label.
\end{IEEEkeywords}}

\maketitle

\IEEEdisplaynontitleabstractindextext

\IEEEpeerreviewmaketitle

\ifCLASSOPTIONcompsoc
\IEEEraisesectionheading{\section{Introduction}\label{sec:introduction}}
\else
\section{Introduction}
\label{sec:introduction}
\fi

\IEEEPARstart{D}{eep} 
\ysq{neural networks (DNNs) have witnessed considerable success in 3D point cloud segmentation in recent years.
Owing to their powerful learning ability, once high-quality annotations are provided, DNNs-based point segmentation methods can achieve remarkable performance. 
However, such strong learning capacity is a double-edged sword, as it can also over-fit label noise and degrade performance if annotations are inaccurate. 
}

\begin{figure*}[ht] 
\centering
\setlength{\tabcolsep}{0.5mm}{
\begin{tabular}{cccc}

\includegraphics[width=0.285\textwidth,page=1]{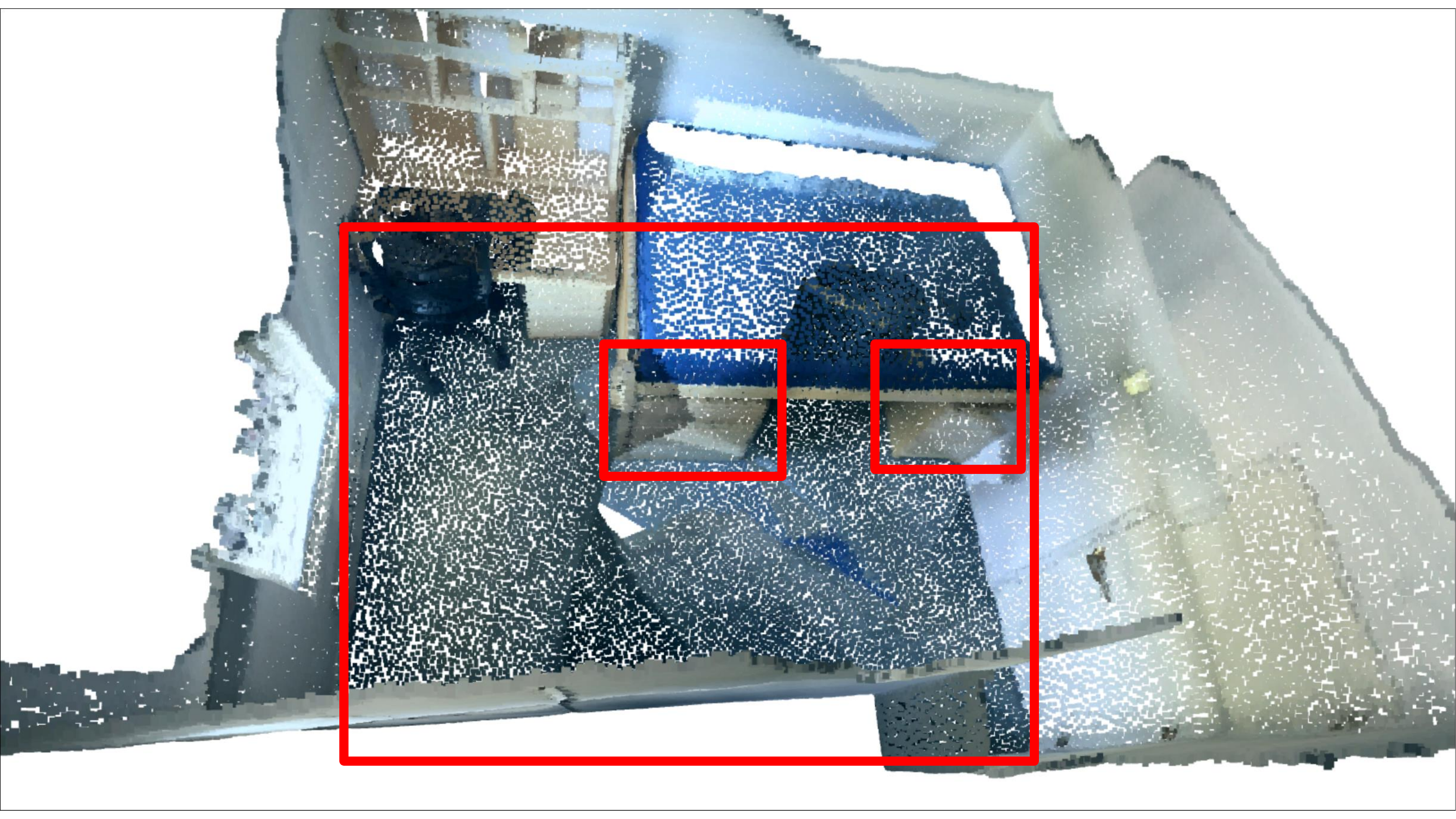} &
\includegraphics[width=0.285\textwidth,page=2]{LaTeX/img/teaser/train_noshuffle_102_oa0.7649055123329163.pdf} &
\includegraphics[width=0.285\textwidth,page=3]{LaTeX/img/teaser/train_noshuffle_102_oa0.7649055123329163.pdf} & \multirow[c]{ 2}{*}[2.3cm]{\includegraphics[width=0.08\textwidth]{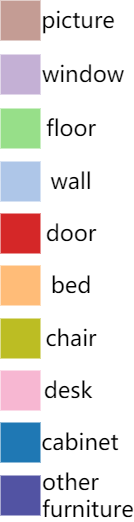}} \\ 
\includegraphics[width=0.290\textwidth,page=1]{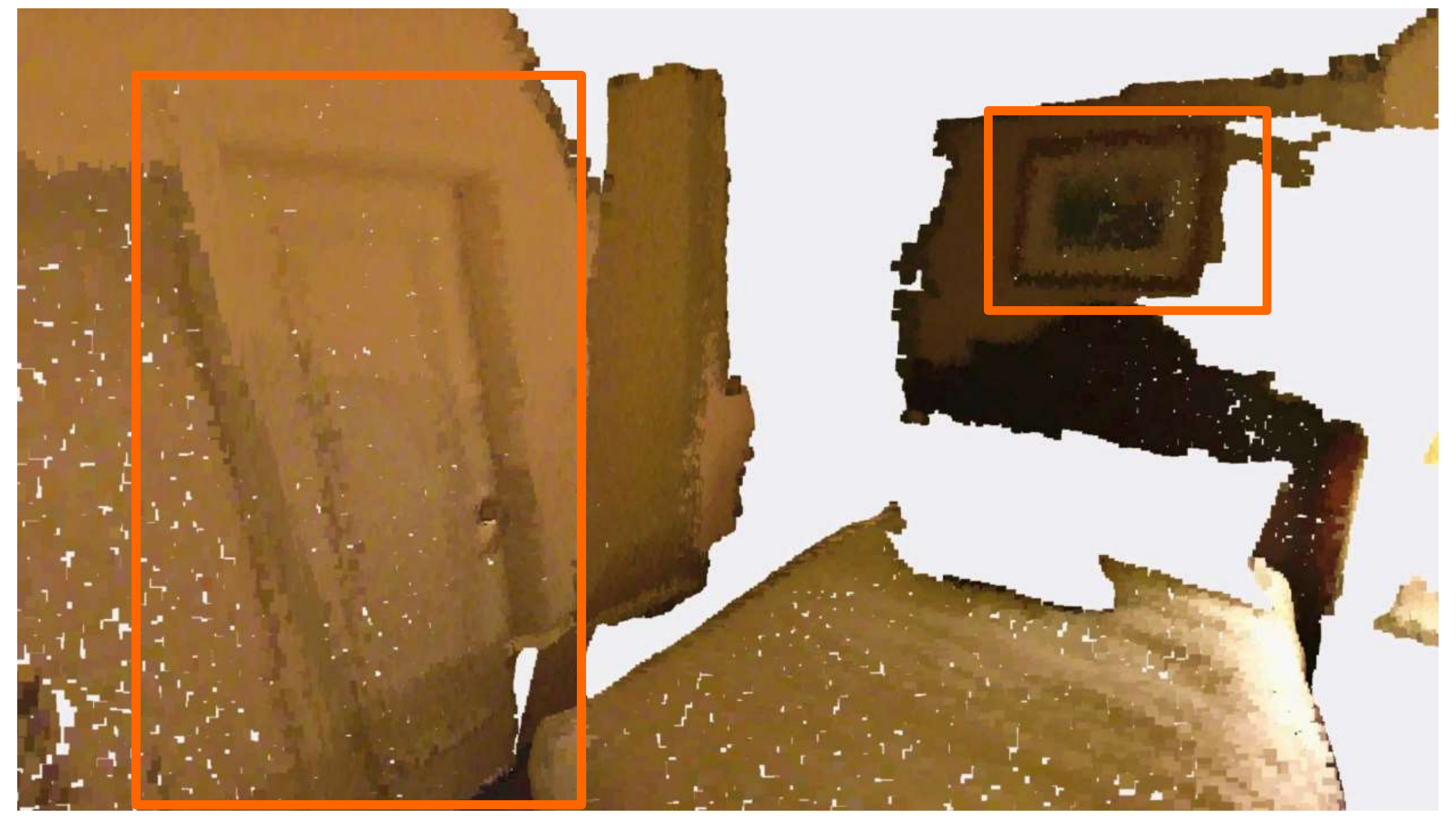} &
\includegraphics[width=0.290\textwidth,page=2]{LaTeX/img/teaser/boundary_tesaer.pdf} &
\includegraphics[width=0.290\textwidth,page=3]{LaTeX/img/teaser/boundary_tesaer.pdf} &  \\
 Input Scenes & Real-World Noisy GT Labels & Predictions of ours & \\

\end{tabular}
}

\caption{Illustration of the instance-level label noise concept in point cloud segmentation. From left to right are the input (noisy instances highlighted red boxes), the manual annotation given by the real-world dataset ScanNetV2, and the prediction of our framework (more in line with the real category \ysq{and the real boundary}). It is noticeable that this popular dataset suffers from label noise, even though it is already a re-labeled version of ScanNet. \ysq{In the first row, the GT label noise is at instance-level, where the floor was mislabeled as a chair. In the second row, we show boundary-level label noise, where we can find inaccurate GT boundaries of door and photo.} Our framework is trained on this noisy dataset but still achieves correct predictions.}
\label{teaser}
\end{figure*}

\ysq{%
In fact, compared to 2D image annotations~\cite{Cordts2016Cityscapes,Tan_2021_TIP_NightCity}, clean labels for 3D data are more difficult to obtain.
It is mainly because: 
1) the point number to annotate is often very massive, \eg, million scale when annotating a typical indoor scene in ScanNetV2~\cite{dai2017scannet}; 
2) the annotation process is inherently more complex, requiring annotators' expertise and additional knowledge, \eg, constantly changing the view, position, and scale to understand the underlying 3D structure. As a result, even the commonly used 3D scene dataset ScanNetV2~\cite{dai2017scannet}, which is already a version after refining the label from the ScanNet, has a large portion of label noise.} \lj{To demonstrate, we divide real-world noisy labels into two main categories: instance-level label noise, where the whole instance has been mislabeled, as shown in the first row of \fref{teaser}, and boundary-level label noise, where the boundary between instances is inaccurately labeled, as shown in the second row of \fref{teaser}.}

\ysq{
Based on the foregoing considerations, research into robust learning with noisy labels for 3D point cloud segmentation is urgently needed.}
\lj{However, to the best of our knowledge, there hasn't been any previous research on point cloud segmentation with noisy labels. The majority of research on learning with noisy labels focuses on image recognition, and  they will fail when directly applied to point cloud segmentation.}
For example, among the most popular methods, sample selection methods \cite{10.5555/3327757.3327944,yu2019does,shen2019learning,song2019selfie,liu2020early} assume that the noise rate of all samples is a known constant value but the noise rates are often unknown and variable. Robust loss function methods \cite{zhang2018generalized,Wang_2019_ICCV} cannot achieve consistent noise robustness to large noise rates. Whereas label correction methods \cite{Reed2015TrainingDN,song2019selfie,ICML2019_UnsupervisedLabelNoise} are designed to correct for image-level label noise, the point cloud segmentation task requires to correct point-level noises. \ysq{Considering that the point labels within each instance are strongly correlated, it is suboptimal to directly apply these methods to each point independently without considering the local correlation.}

\ysq{We make the first attempt to empower the point cloud segmentation model with resistance to different types of real-world annotation noise.
By presenting a novel point noise-adaptive learning (PNAL) framework, we take the lead in solving the instance-level point cloud label noise problem. We also design a variant of PNAL, called PNAL-boundary, to resist boundary-level label noise in point cloud and combine both PNAL and PNAL-boundary to solve the mixture of instance-level and boundary-level noises.}
Specifically, to cope with unknown, possibly heavy, and varying noise rates, we designed a point-level confidence selection mechanism, which obtains reliable labels based on the historical predictions of each point without requiring a known noise rate. Next, in order to fully utilize the local correlation among labels, we propose a label correction process performed at the cluster-level. This is done by the proposed voting strategy that tries to merge reliable labels from relevant points to provide the best possible label for each point cluster, with a computationally efficient implementation. 
\hsf{As a result, the proposed PNAL framework is able to clean point labels concurrently by gradually investigating the majority of points in the complete dataset. }

\hsf{However, applying PNAL directly to data with boundary-level label noise can result in incorrect label correction on interior points. So we extend it to PNAL-boundary, which is able to correct the point labels near the underlying boundary while keeping clean labels of inner points intact. As the groundtruth border is unknown, a progressive boundary cleaning strategy is introduced to PNAL-boundary for updating boundary regions iteratively with predicted point labels.}

\lj{Our method has been qualitatively and quantitatively validated on both synthetic noisy label dataset generated from the standard large-scale 3D
indoor spaces (S3DIS)~\cite{armeni_cvpr16} and real-world noisy dataset ScanNetV2 \cite{dai2017scannet}. For a more rigorous evaluation, we refine the validation set of ScanNetV2 by manually correcting the noisy labels. Experiments on 
instance-level, boundary-level, and mixture noisy settings show that our framework achieves better performance and time efficiency than adapting existing noisy label correction works (SCE~\cite{Wang_2019_ICCV}, GCE~\cite{zhang2018generalized}, SELFIE~\cite{song2019selfie}) to the 3D point cloud segmentation task, and is even comparable to the upper bound trained on completely clean data. }

To summarize, our contributions are fourfold. 
\begin{itemize}
\setlength{\itemsep}{0pt}
\setlength{\parsep}{0pt}
\setlength{\parskip}{0pt}
\item To the best of our knowledge, this is the first work investigating noisy labels on point cloud data, which has a wide and urgent need for 3D applications where the volume of data is growing drastically.
\item A novel noise-rate blind framework PNAL is proposed to handle spatially variant noise rates in point cloud. It consists of point-level confidence selection, cluster-level label correction with voting mechanism, and can be easily applied to different network architectures. 

\item \lj{In addition to addressing instance-level label noise with the standard PNAL, we propose a variant, i.e., PNAL-boundary, which can deal with boundary-level label noise using a novel progressive boundary cleaning strategy. And we further combine PNAL and PNAL-boundary to solve a mixture of both instance-level and boundary-level label noise.}

\item We re-labeled the validation set of ScanNetV2 by correcting noises and will make it public to facilitate both point cloud segmentation and noise label learning.

\end{itemize}

\section{Related Work}
\label{sec:related}

\subsection{Point Cloud Segmentation.}
Given a point cloud, the task of semantic segmentation is to divide it into subsets based on the semantic meaning of the points. Among the related works, point-based networks have recently gained more and more attention. These methods act directly on disordered, irregular, and unstructured point clouds, so that directly applying standard CNNs is not feasible. For this reason, the pioneering work PointNet \cite{qi2017pointnet} was proposed to learn per-point features using cascaded multi-layer perceptrons. It uses cuboidal blocks of fixed arbitrary size to cut the rooms in the scene datasets into blocks when processing datasets. Inspired by PointNet, a series of point-based networks have been recently proposed and widely used today~\cite{Ye2021MetaPUAA,ye20213d,han2021exemplarbased,gong2021boundary}. In general, following the practice of PointNet, these approaches use octrees \cite{tatarchenko2017octree}, kd-trees \cite{klokov2017escape} or clustering \cite{landrieu2018large} to decompose room scenes into blocks. \lj{Although these methods have achieved great success in point cloud segmentation, they are based on the clean label assumptions. Once noisy labels exist, their strong learning capacity will make them over-fit inaccurate annotations and thus cause performance degradation.}

\subsection{Image Recognition with Noisy Labels.}
\lj{Most existing noisy label correction methods are designed for the image recognition task, which can be broadly classified into four categories. }
First, methods with noise-robust layers \cite{7298885,Goldberger2017TrainingDN,7472164} are intended to model a label transition matrix $\mathrm{T}$. Based on the estimated $\mathrm{T}$, they adjust the output of the network to a more confident label. However, such methods assume a strong correlation between certain labels, which limits the method usages. 
The second approach is to design a loss that is robust to noisy labels
\cite{ghosh2017robust,DBLP:conf/iclr/LyuT20}, generalized cross entropy (GCE)\cite{zhang2018generalized}, symmetric cross entropy (SCE)\cite{Wang_2019_ICCV}
, which are popular and easily adaptable to existing architectures. They are originally proposed for classification, but most of them are applicable to segmentation if used pixel-wisely. A recent work \cite{yang2020lncis} proposes to apply the reverse cross entropy~\cite{Wang_2019_ICCV} for the foreground-instance sub-task and a standard cross entropy (CE) loss for the foreground-background in instance segmentation. A major limitation is that they cannot handle heavy noisy labels. 
Third, loss adjustment methods \cite{Reed2015TrainingDN,NIPS2017_2f37d101,NEURIPS2018_ad554d8c,ICML2019_UnsupervisedLabelNoise,song2019selfie,zhang2020characterizing,wang2020noise} reduce the negative impact of noisy labels by adjusting the loss of all training samples. In general, while these methods are fully exploring the training data, they take the risks of false correction. 
Among them, \cite{zhang2020characterizing,wang2020noise} are the latest methods that try to solve the noisy label problem on binary segmentation. 
Forth, sample selection methods \cite{10.5555/3327757.3327944,yu2019does,shen2019learning,song2019selfie,liu2020early,pnal2021} select true-labeled samples from noisy data to avoid false correction, but they take the risk of discarding usable samples and require either a true noise rate or a fully clean validation set. 

\lj{Despite their success in handling noisy labels for image recognition, applying the above-mentioned methods to point cloud segmentation will fail because point cloud data has a different representation and spatially-variant unknown noise rates. We take the first step to address label noise in point cloud segmentation.} Our approach is an extension and hybrid of sample selection and loss correction methods. Unlike sample selection methods, no noise rate is required in our method. And unlike loss correction methods that correct all samples, we correct labels of point clusters according to confidence considering their locally similar region.

\subsection{Edge Detection and Image Segmentation with Noisy Lables.}
Besides solving noisy labels in image recognition, some methods also consider inaccurate boundary labels for edge detection and image segmentation tasks. For edge detection, the boundary-level label noise problem that occurs frequently for professional annotation is even more severe for crowd-sourcing labels, as first reported by \cite{yu2018seal}. They thus propose an edge detector that helps to align the noisy edge annotations with the true boundaries. 
Based on it, \cite{AcunaCVPR19STEAL} propose a new thinning layer and NMS loss on edge detection networks to produce thin and precise edges. 
The major limitation of these frameworks is that they assume the noise rate is small. Also, they cannot deal with instance-level label noise, and are limited to the domain of 2D images. More importantly, such approaches only perform edge detection instead of semantic segmentation, which makes them not applicable to the task of point cloud segmentation on their own.

\ysq{Boundary label noise also occurs in the domain of medical image segmentation and binary object detection~\cite{zhang2018deep,PMD:2020,GSD:2021}, and several attempts~\cite{zhu2019pick,xue2020cas,zhang2020robust,zhang2020characterizing,shi2021distilling,li2021superpixel} have been made to tackle this problem by extending different learning methodology with noisy labels. 
However, they only focus on binary segmentation instead of multi-class segmentation in this task, and most of them require a true noise rate. Our proposed framework can handle multi-class scenario without the need of noise rate.}

\section{Problem Description}
\label{sec:taskdef}
\ysq{In this section, we define instance-level label noise and boundary label noise problem in point cloud segmentation, according to our observations on popular real-world datasets.}

We define the task of multi-class point cloud semantic segmentation with noisy labels. 
Formally, we denote point cloud data as $\mathbf{X} \in \mathcal{R}^{N \times C}$ of $N$ points with $C$ features of coordinates and RGB values possibly, and its semantic label as $\mathbf{Y}$, and $M$ as the class number. Our target is to train a model $\mathbf{f}_{\theta}(\mathbf{X})$ robust to both instance-level and boundary-level label noise in the training set.

\subsection{\ysq{Instance-level Label Noise Problem}}
\ysq{For real-world datasets such as ScanNet~\cite{dai2017scannet} and ScanNetV2, instance-level mislabeling can be frequently observed.}
According to our observations, while the semantic label of an instance might be wrong, we seldom encounter incorrect instance partitioning. For example, a table may be labeled as a sofa, but we seldom see a table that is partially labeled as a sofa. This implies that our method should correct the noise at the instance level.  However, the ground truth instance information may not available, and predicting object instances itself is a challenging task. 
Alternatively, we create a cluster based noise correction method, where each cluster consists of a local patch of points. The main assumption is that points in a cluster are considered to belong to the same instance.

\subsection{\ysq{Boundary-level Label Noise Problem}}
\ysq{Besides the instance-level label noise problem, another challenging issue is the problem of label noise near the true boundary between objects. 
For example, as shown in the second row of \fref{teaser}, the points whose real category is wall may be mislabeled as a door, when its position is close to the true boundary between the wall and the door. 
For the crowd-sourced ScanNetV2 dataset with annotations, where quality control is a challenge, the boundary label noise spreads across the entire dataset. According to our experiments, the boundary noise labels can have a serious impact on the model performance, even though they represent a very small fraction of the overall number of point clouds.
Generally, there are significant differences in geometry and color distribution between the points located near the ground truth boundary edges of adjacent objects. It enables the granularity of clustering methods to be coarse-grained at the inner areas of the objects and become fine when closer to the real boundaries, and can thus be naturally used to refine the coarse boundary labels.  %
This inspired us to design a cluster-based boundary noisy label correction approach. } 

\subsection{Point Cluster}
\lj{As analyzed above, both instance-level and boundary-level label noise indicate it is better to be corrected based on clusters instead of points.} We therefore generate point clusters with off-the-shelf clustering algorithms. In experiments, we mainly use the DBSCAN~\cite{ester1996density}. 
DBSCAN is a density-based clustering algorithm that does not require the specification of the cluster number in the data, unlike k-means. DBSCAN can find arbitrarily shaped clusters, and this characteristic makes DBSCAN very suitable for LiDAR point cloud data. 
In the experiment we also tried another clustering method GMM. \ysq{Note that the feature used in the clustering methods includes not only point positions, but also the corresponding RGB color attribute.}
During training, we correct the label on a cluster level. Based on the experiments described in Sec.~\ref{subsec:abl2_vsEpsandGMM}, we demonstrate that our method is robust to the granularity of cluster to a certain range and insensitive to clustering methods. We denote the point cloud and its label in a cluster $\mathbf{C}_{i}$ as $(X_{\mathbf{C}_{i}}, Y_{\mathbf{C}_{i}}), 1 \leq i \leq k$, and $k$ is the number of clusters.

\section{Methodology}
\label{sec:methodology}

\ysq{In this section, for the proposed PNAL for instance-level label noise and the PNAL-boundary for boundary label noise, we first introduce the shared pipeline of the two, and then introduce their different part respectively.}

\subsection{Pipeline Overview}

\begin{figure*}[ht]
\includegraphics[width=\textwidth]{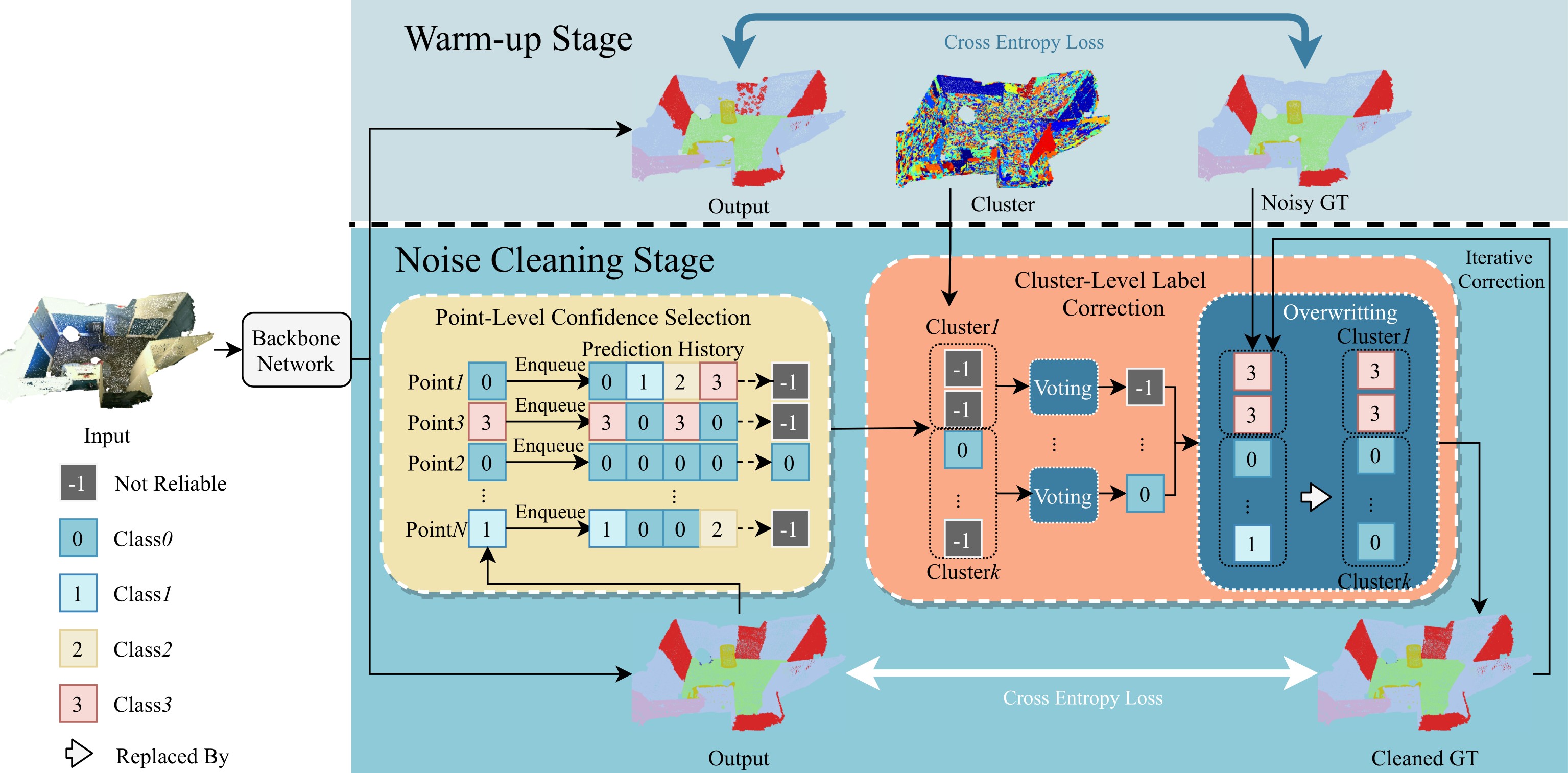}
\caption{System pipeline \ysq{of the proposed PNAL for instance-level noise}. In the warm-up stage, the network is updated with CE as usual.
In the noise cleaning stage, we enqueue the output to the prediction history and point wisely perform confidence selection to get reliable labels. With these results, we do voting at cluster level, then correct the original noisy GT or the previously cleaned GT. Finally the obtained cleaned GT guides the network update.
}
\label{Fig:pip}
\end{figure*}

\begin{figure}[ht]
\includegraphics[width=0.45\textwidth]{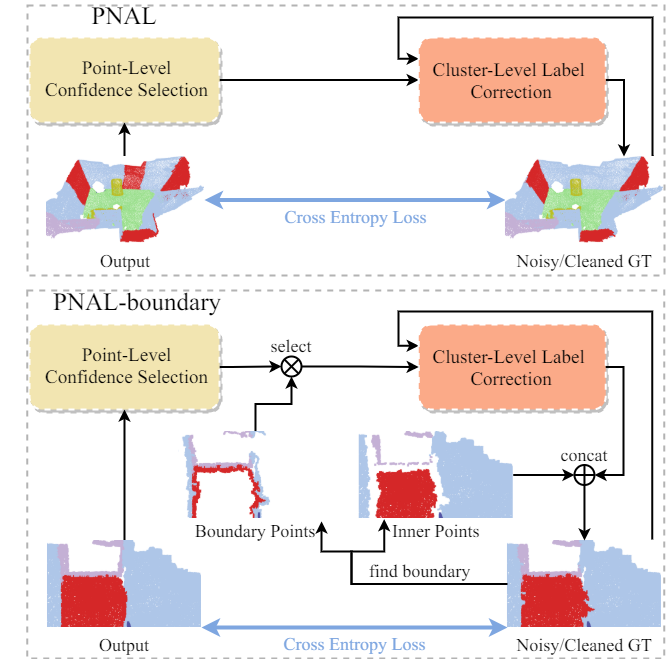}
\caption{\ysq{Illustration diagram to show the training difference between the PNAL and PNAL-boundary on noise cleaning stage.}
}
\label{Fig:pipdiffer}
\end{figure}

\ysq{Our training procedure for both PNAL and PNAL-boundary is divided into two stages, a warm-up stage introduced in Sec.~\ref{sec:warmup} and a noise-cleaning stage introduced in Sec.~\ref{sec:selection}, Sec.~\ref{sec:correction} and Sec.~\ref{sec:pnalboundary}. The warm-up stage is the same for both frameworks, which is illustrated in the upper (light blue) part in Fig.~\ref{Fig:pip}.} We first train the network on all samples by default manner for $E_{warm-up}$ epochs, where $E_{warm-up}$ denotes the number of epochs for this stage. The warm-up stage motivates the network to learn the easy data, which are largely consistently labeled correct data.
 
In the PNAL pipeline, we further train the network with our proposed noise-cleaning stage, as shown in the lower (sky blue) region in Fig.~\ref{Fig:pip}. The main idea is to identify the potential noisy label based on the behavior of the network prediction and update the selected data label to a more reliable label. From the observation that the ground truth label is often corrupted on an instance-level, we encourage to correct data labels in a group-wise manner. That we propose to first cluster point cloud into small patches and then update point label patch-by-patch. Point cloud in one patch is set to the same semantic label. Besides, we predict the new label of each cluster through a voting strategy. 

\ysq{In the PNAL-boundary, pipeline for its noise cleaning stage is slightly different from the PNAL's, as illustrated in Fig.~\ref{Fig:pipdiffer}. To progressively correct the coarsely annotated boundary labels while protecting the correct non-boundary labels, we propose a new progressive boundary cleaning strategy, as introduced in Sec.~\ref{sec:pnalboundary}.}

\subsection{Warm-Up Stage}\label{sec:warmup}
According to the study of memorization effects \cite{arpit2017closer}, deep neural networks are prone to learn clean, easy samples first, and then other noisy samples gradually, even in the presence of noisy labels. Therefore, in the warm-up stage, we adapt no strategies and train the network with a common cross-entropy loss. The detailed formula of $\operatorname{Loss}_{warm-up}$ is:
\begin{equation} \label{eq:1}
\operatorname{Loss}_{warm-up}=-\frac{1}{B} \sum_{i=1}^{B} \sum_{m=0}^{M} q\left(m \mid \mathbf{X}_{i}\right) \log p\left(m \mid \mathbf{X}_{i}\right),
\end{equation}
where $B$ is the number of samples inside a mini-batch, $ p\left(m \mid \mathbf{X}_{i}\right)$ denotes classification confidence of each class $m \in \{1,...,M\}$, and $q\left(m \mid \mathbf{X}_{i}\right) \in {\{0,1\}}^{N \times M}$ is one-hot encoded label. 

The warm-up stage power the network with easily learned data samples. However, if trained with a large number of epochs for the warm-up stage, the network will tend to fit noisy labels. Let $E_{clean}$ denotes the number of epochs for the noise-cleaning stage. In noise-cleaning stage of training, remarkably, no hyperparameters such as noise rate are required. 
We observe that, the precision of both the replaced labels and the correctly fixed labels increases gradually as the training step progresses, and label replacement eventually expands to almost the entire training set, as analysized in \ref{subsec:corrprocess}.
Moreover, through our analysis in \ref{subsec:abl1_Ewarmupvsnoiserate} on $E_{warm-up}$, too much warm-up causes the model fitting more noisy data, which affects the performance when the noise rate is large. Meanwhile, the optimal $E_{warm-up}$ setting is not sensitive to the noise rate. At different noise rates, we can derive the following relationship between the optimal $E_{warm-up}$ and $E_{clean}$,

\begin{equation} \label{eq:2}
E_{warm-up}=\frac{1}{5}E_{clean}.
\end{equation}

\subsection{\ysq{Noise Cleaning Stage}}

\subsubsection{Point-Level Confidence Selection}\label{sec:selection}

In this part, we aim to select reliable samples from each mini-batch and obtain the confidential labels for these samples that can be corrected with high probability.

Previous studies of sample selection \cite{song2019selfie,10.5555/3327757.3327944} adopt the widely used loss-based separation which tries to construct clean set $\mathcal{C}$ by selecting $(1-\tau) \times 100\%$ of low-loss instances, where $\tau$ is the noise rate. While significant improvements have been achieved in these works, such an approach faces two key problems. First, it is not appropriate to assume that the noise rate is available and constant on the 3D point cloud. Second, it would exclude $\mathcal{C}_{h}$ which tends to produce high losses, making it difficult for hard cleans to participate in the network update. To address the above weaknesses of the loss-based sample selection methods, instead of selecting clean samples, we propose to directly select the reliable samples without knowing the noise rate.

Inspired by the bootstrapping \cite{Reed2015TrainingDN}, we develop a confidence point selection method for dynamically selecting reliable samples and target labels without requiring explicit noise rate. Instead of using predicted class probabilities to generate regression targets, a new criterion, inspired by SELFIE \cite{song2019selfie}, is designed based on history prediction. In detail, a sample with consistently predicted label is regarded as the reliable sample, and its most frequently predicted label is the reliable label, as defined below:

\textup{Definition 1.} A point $x_{{n}}$ is a reliable sample if the predictive confidence $F(x_{{n}} ; q)$ satisfies $F(x_{{n}} ; q) \geq \sigma (0 \leq \sigma \leq 1)$.
The predictive confidence is defined as 
\begin{equation} \label{eq:2}
F(x ; q)=(1 / \lambda) \text { entropy }(P(m \mid x ; q))\frysq{,}
\end{equation}
where $\lambda = - \log (\frac{1}{M})$ is a normalization term for normalizing to $[0,1]$.
We denote the predicted label of the sample point $x$ at time $t$ as
$\hat{m}_{t}=\mathbf{f}_{{\theta}_{t}}(x)$. Then, the label history of the sample $x$ that stores the predicted labels of the previous $q \leq E_{warm-up}$ times is $H_{x}(q)=\left\{\hat{m}_{t_{1}}, \ldots, \hat{m}_{t_{q}}\right\}$, where $q$ is history length. 
Next, as given in Eq. \ref{eq:3}, $P(m \mid x ; q)$ is the probability of the label $m \in\{1,\ldots, M\}$ estimated as the label of the sample point $x$. 
\begin{equation} \label{eq:3}
P(m \mid x ; q)=\frac{\sum_{\hat{m} \in H_{x}(q)}[\hat{m}=m]}{\left|H_{x}(q)\right|}\frysq{,}
\end{equation}
where $[\cdot]$ is the Iverson bracket notation. Then we denote the set of reliable samples as $\mathbf{X}_{reliable}$.
Finally, we define the reliable label as $m^*_{{n}}$ where
$$
m^*_{{n}}=\arg \max_m P(m \mid x_{{n}} ; q).
$$

Specifically, our method differs from SELFIE in three significant ways. 
First, our method is noise-rate blind, to handle spatially variant noise-rate problem in point cloud. 
Second, our confidence selection mechanism is point-level, and our label correction process is done on cluster level with the help of a novel voting strategy, in order to consider local relationship between point labels. Note that the reliable label in selection stage may not be the label won for correction process.
Last but not least, our detailed implementation has to be much more computationally efficient, to allow for point-level selection and cluster-level correction.

\subsubsection{Cluster-Level Label Correction}\label{sec:correction}

The label correction process is to selectively replace the label with the best possible label in locally similar regions. Ideally, the local similarity region is defined by the ground-truth instance. However, since instance labels may not be available in practice, we use cluster as an alternative. From each cluster $\mathbf{C}_{i} (1 \leq i \leq k)$, as given in Eq. \ref{eq:4}, the ones containing reliable samples will be selected for the label correction in subsequent steps.

\begin{equation} \label{eq:4}
\{ \mathbf{C}_{i} \mid \exists x_{{n}}: x_{{n}} \in \mathbf{X}_{reliable} \land x_{{n}} \in \mathbf{X}_{\mathbf{C}_{i}} \}\frysq{,}
\end{equation}
where $n \in\{1,\ldots, N\}$. 
Next, for each of these clusters, \eg $\mathbf{C}_{i^*}$, we will replace the label $\mathbf{Y}_{\mathbf{C}_{i^*}}$ with a best label locally. The goal label can be found by the proposed voting strategy according to the overall label occurrences within the cluster. We use $occ^{m}_{i^*}$ to denote the occurrence number of reliable samples with $m^*_{{n}}=m$.

\begin{equation}
\begin{aligned}
occ^{m}_{i^*} {} & = \sum_{n=1}^N [m^*_{{n}}=m \land x_{{n}} \in \mathbf{X}_{\mathbf{C}_{i^*}}] \\
        & =\sum_{n=1}^N [m^*_{{n}}=m] [x_{{n}} \in \mathbf{X}_{\mathbf{C}_{i^*}}]\frysq{,}
\end{aligned}
\end{equation}
where $[\cdot]$ is the Iverson bracket.
Then, occurrences for each class are formed as a vector $occs_{i^*} = \left( occ^1_{i^*}, ..., occ^{M}_{i^*} \right)$, and we can find the top occurrence by $occs^{top}_{i^*} = \max_{1\leq m \leq M} occs^{m}_{i^*}$ . A winner label is randomly chosen from $ \{m \mid occ^{m}_{i^*} \geq \frac{ occs^{top}_{i^*} }{\gamma} \}$ to overwrite the label of this cluster.
Note that, in a special case of $\gamma=1$, the winner label in this cluster will be from the top reliable labels $\{m \mid occ^{m}_{i^*} = { occs^{top}_{i^*} } \}$.
According to our ablation study, $\gamma=4$ achieves the best performance. Note that the original label may not be mislabeled or different from the winner label. And the labels may be repeatedly overwritten during the training process.

\subsection{\ysq{Instance-level Label Cleaning}}

\ysq{In the noise cleaning pipeline of PNAL framework, we first feed the latest output of the backbone network into the point-level confidence selection, where each points will enqueue into the prediction history and marked as reliable or unreliable. Then, for each points equally, we perform cluster-level label correction, and iteratively replace the ground truth labels.
Finally, we update the backbone network with these replaced labels by a cross-entropy loss, and the samples whose label has never been replaced are not involved in the gradient calculation.}

\subsection{\ysq{Boundary-level Label Cleaning}}\label{sec:pnalboundary}

\ysq{The noise cleaning pipeline for PNAL-boundary is slightly different from PNAL. The former is guided by the prior of the boundary label noise.}
\ysq{To describe the boundary label noise, we first define the boundary points. The boundary points are the points at the real boundary between objects of different classes. When segmentation annotations are really clean, the boundary points can be obtained by finding the points that more than one classes exist in its nearest neighbors' labels. 
According to our observation on real-world datasets, the label noise often occurs at a certain range near the true boundary, i.e., erosed or dilated labels can be found among the points $\mathbf{X}_{boundary}$ which belongs to the nearest neighbors of the true boundary points (including boundary points themselves). 
The erosion or dilation pattern for label noise is highly random, and the point's noise level decreases as its distance to the boundaries increases.
Thus, different from symmetrical or asymmetrical pattern of instance-level label noise, a specific morphological algorithm should be designed to model the boundary label noise, as described in Sec.~\ref{boundarynoise_algorithm}.
Guided by the observation, our correction method should focus on cleaning the labels of the true boundary points and the points near the true boundary, while preserving the non-boundary (inner) point labels. 
} %

As discussed earlier, in an ideal situation, we only correct point labels near the true boundary and protect the inner labels. However, from the given ground-truth label which is noisy near the boundary, we cannot find the true boundary. Thus, we propose a progressive boundary correction strategy. The key idea is to adopt label correction for the most recent boundary points extracted from the latest label which is progressively updated as training goes, while protecting the label for inner points. 
In detail, in the first epoch of the noise cleaning stage, we obtain the set of boundary and neighbor points $\mathbf{X}_{boundary}'$ by the $k$ -nearest neighbors based on the initial noisy ground truth, and the rest points are inner points. According to our ablation study, $k=20$ achieves the best performance. Note that $\mathbf{X}_{boundary}' \neq \mathbf{X}_{boundary}$, because the ground truth label in our training set is noisy.
In the following epochs, unlike the noise cleaning stage of PNAL, we adopt different strategies for boundary-neighbor points and inner points. First, we perform point-level confidence selection for all points. Second, for boundary and neighbor points $\mathbf{X}_{boundary}'$, we perform cluster-level label correction within itself, i.e., the reliable samples are selected from boundary and neighbor points:
\begin{equation} \label{eq:7}
\{ \mathbf{C}_{i} \mid \exists x_{{n}}: x_{{n}} \in \mathbf{X}_{boundary}' \land x_{{n}} \in \mathbf{X}_{reliable} \land x_{{n}} \in \mathbf{X}_{\mathbf{C}_{i}} \}\frysq{.}
\end{equation}
For inner points $\mathbf{X} - \mathbf{X}_{boundary}'$, we stop replacing its ground truth label. Then, the ground truth is updated by the labels of the corrected boundary points and all of the inner points. Finally, the network will be updated by the replaced labels. The gradient calculation excludes boundary points whose labels have never been replaced.
For each epoch, in addition to correcting the ground truth label iteratively, we also obtain the new boundary and neighbor points according to the latest cleaned ground truth.
The number of neighbors $k$ to encountered within each correction is limited to a certain range, in order to minimize incorrect modifications to the real inner points. According to our ablation, $k=20$ achieves the best performance.

\noindent\textbf{Connection with Instance-level Cleaning. }\frysq{ We note that the instance-level and boundary-level cleaning pipeline have the following connections. First, they share the two-stage learning pipeline for training, i.e., a warm-up followed by a noise cleaning stage. Second, they both perform Point-Level Confidence Selection to select reliable samples and Cluster-Level Label Correction for label correction in the noise cleaning stage. Moreover, it will be shown in following experiment (i.e., Table~\ref{Tab:s3disboundary}) that the PNAL framework also works for boundary-level label noise to some extent.}

\section{Datasets and Noise Settings}
\label{sec:dataandnoise}
To demonstrate the effectiveness of our proposed method, we conduct experiments on two datasets, ScanNetV2 \cite{dai2017scannet} and S3DIS \cite{armeni_cvpr16}. ScanNetV2 is a popular 3D real-world dataset with label noise. S3DIS is a commonly used scene dataset with much cleaner labels, which can be considered as clean data. There we can artificially build noisy datasets from S3DIS with various noise settings.

\subsection{ScanNetV2} 
The ScanNetV2 3D segmentation dataset contains $1,513$ annotated rooms with $21$ semantic elements in total. According to the scan annotation pipeline for ScanNetV2 \cite{dai2017scannet,smartscenes}, a normal-based graph cut image segmentation method \cite{felzenszwalb2004efficient} is first utilized to get a basic, pre-segmentation. These provide a reliable reference to an object instance, however, the class label can be carelessly mislabeled in practice.
Also, since the rooms are distributed to different annotators, inconsistent labels can be found even for the same object with different placements in the same scene.
These observations match our assumption about the noise pattern in point cloud segmentation that mislabeling occurs at the object-instance-level. 
And we noticed that the noisy label problem also exists even in its validation sets. Since this issue has never been mentioned in other studies, we manually correct such noise labels of all scenes from the validation sets of ScanNetV2 for more accurate evaluation.
Note that we did not perform evaluation on the benchmark test split, due to its unknown noise rate and unavailable annotation.

\subsection{S3DIS} 
S3DIS contains point clouds of $272$ rooms in $6$ large-scale indoor scenes in three buildings with $12$ semantic elements. 
The instance label is borrowed from \cite{pham-jsis3d-cvpr19} which are manually annotated. 
Compared to ScannetV2, the S3DIS dataset has a much smaller amount of scenes, less scene complexity, and fewer classes, and the errors occurring in class labels are clearly less than the former. Therefore, we treat the S3DIS dataset as a completely clean dataset, i.e., the noise rate is $0$.

\subsubsection{\ysq{Instance-level label noise}} 
We generate an instance-level noisy dataset from S3DIS by randomly changing the point label at an object-instance level, guided by our noise pattern assumption.
Following previous work on image classification with noisy data, we model the noisy dataset with two noise types: symmetric and asymmetric. For symmetric noise, the point label is modified to other labels with equal probability at instance level. 
Also, we found that some class pairs are easily mislabeled as each other
in the real-world noisy ScanNetV2, such as door and wall, while some pairs are not that confusable, such as wall and desk. Based on this, we create asymmetric noisy S3DIS datasets, mimicking the way in real-world. 
In particular, we identify the easily misclassified label pairs, including door-wall, board-window, sofa-chair, and randomly flip the label inside each label pair with a probability of $\tau_{pair}$. To note that ours setting is different from previous work on image noisy labels. On point cloud data with only $12$ semantic classes, it is inappropriate and unrealistic, to define another confusable class for all classes. Therefore, for classes without pairs, we follow the symmetrical noise setting, to achieve a specified value of the overall noise rate $\tau$. We will show results in the supplemental file following the asymmetric noise setting in previous works on image.

\subsubsection{\ysq{Boundary-level label noise}} \label{boundarynoise_algorithm}
\ysq{
To quantify the noise level that PNAL-boundary can handle, we also simulate the boundary label noisy dataset from S3DIS. Inspired by our observations on real-world noisy datasets in \sref{sec:pnalboundary}, we model the label noise on semantic boundary as an erosion or dilation operation on labels. It is highly random, and the point's noise level decreases as its distance to the boundaries increases. In detail, to produce datasets with various noise levels, we randomly choose a ratio $\alpha$ of scans from the training data to perform morphological noise generation shown in Algorithm~\ref{alg:boundary} with noise level controlled by $\beta$. 
}

\begin{algorithm}[hbt!]  \label{alg:boundary}
\SetAlgoLined

\caption{Morphological label noise generation} 
\KwIn{ Point\_Positions $\{\mathrm{X} \in \mathbb{R}^{N \times 3 } \}$, Label $\{\mathrm{Y}\}$, Sample\_Noise\_level $\{\beta\}$}
\KwOut{ Morphological\_Noisy\_Label $\{\mathrm{Y'}\}$}

initialize $\mathrm{Y'} \leftarrow \mathrm{Y}$

\While{ $\textrm{count\_point\#(} \mathrm{Y'} \neq \mathrm{Y}) < \textrm{threshold\_point\#(}\beta)$}{ 
    \tcc{loop if the number of noisy labels is fewer than a threshold controlled by $\beta$}

    $y \leftarrow$ random\_sample( get\_unique\_elements($\mathrm{Y}$) )

    \tcc{to avoid class imbalance}

    $i \leftarrow$ random\_sample ( get\_index($\mathrm{Y} == y$))

    nearest\_neighbors\_indexs $\{nn\} \leftarrow$ knnsearch ( $\mathrm{X}_i, \mathrm{X}, \mathrm{k}=80$ neighbors)

    \If{any ($\mathrm{Y}_{nn} \neq \mathrm{Y}_i$)}
    {
        \tcc{point $i$ is at the boundary, so do label noise to neighbors}

        distance\_point\_to\_neighbor $\{\mathrm{dneighbor}\} \leftarrow$ distances $\left(\mathrm{X}_i, \mathrm{X}_{nn}\right)$

        average\_distance $\{\mathrm{davg}\} \leftarrow$ mean ($\mathrm{dneighbor}$ )

        mask\_to\_flip\_label $\{\mathrm{mask}\} \leftarrow $ distance\_related\_random\_sample ($\mathrm{dneighbor},\mathrm{davg},\beta$)

        \tcc{The closer a neighbor point is to point $i$, the more likely it is to do label flipping.}

        $\mathrm{Y'}_{nn}[mask] \leftarrow \mathrm{Y}_i$

    }

}
return $\mathrm{Y'}$

\end{algorithm}

\section{Experiments}

\subsection{Implementation Details}
The DBSCAN algorithm is used for point cloud segmentation in this study, if not specifically stated, with $\varepsilon=0.018$. 
For the real-world noisy dataset ScanNetV2 and artificially created noisy dataset S3DIS, room scenes are divided into room blocks of size $1.0 \times m$ and stride $0.5 \times m$. We random sample $4096$ points for each room block during training. 
We report the results in terms of overall accuracy (OA) and mean intersection over union (mIoU) with DGCNN~\cite{dgcnn}, \frysq{PointNet2}~\cite{qi2017pointnetplusplus}, and SparseConvNet~\cite{3DSemanticSegmentationWithSubmanifoldSparseConvNet} as backbones.
Without special notation, all experiments are conducted with DGCNN as backbone. 
For symmetric noise, we conduct experiments on noise rates $\tau \in \{20\%,40\%,60\%,80\%\}$. 
For asymmetric noise, we test on a large noise rate $\tau = 60\%, \tau_{pair} = 40\%$. 
\ysq{For boundary noise, we conduct experiments on dataset noise rates $\alpha \in \{0.5,1.0\}$ and sample noise rate $\beta \in \{0.3,0.5,0.7\}$.}
All the results on S3DIS are tested on the clean 6th-Area. We train a total of 30 epochs,  including the warm-up stage and clean-noise stage. The history length is set to 4.

\subsection{Baselines}
Note that we are the first handling noisy label on point cloud segmentation. We try our best to adapt previous works and create the following baselines: CE, GCE\cite{zhang2018generalized}, SCE\cite{Wang_2019_ICCV} and SELFIE\cite{song2019selfie}. 
The first three methods can be naturally applied to the point cloud segmentation as point level guidance.
To adopt SELFIE, we apply the original implementation of image-level SELFIE point-wisely
. We experimentally find that the optimal warm-up threshold is $5$. The other settings are the same as in their paper.

\subsection{Performance Comparison on S3DIS}

\subsubsection{Instance-level Label Noise}

\begin{table*}[ht]
\centering
\caption{OA Comparison of different methods on artificially created noisy S3DIS. The tops with different backbones are shown in bold.}
\label{Tab:s3dis}
\setlength{\tabcolsep}{3.3mm}{
\begin{tabular}{@{}c|c|cccc|c|c|c@{}}
\toprule
\multirow{2}{*}{Methods} & \multirow{2}{*}{0\%} & \multicolumn{4}{c|}{Symmetric Noise ($\tau$)}                                  & Mixed Asymmetric Noise & \multicolumn{2}{c}{Pure Asymmetric Noise ($\tau$)} \\ \cmidrule(l){3-6} \cmidrule(l){7-9} 
                         &                      & 20\%            & 40\%            & 60\%            & 80\%            & $\tau_{pair}=40\%$,$\tau=60\%$     & 20\%                   & 40\%        \\ \midrule 
DGCNN\cite{dgcnn}+CE                 & \textbf{0.8692}      & 0.7506          & 0.6732          & 0.6390          & 0.5060          & 0.5634          & 0.8288                 & 0.7513  \\
DGCNN\cite{dgcnn}+SCE\cite{Wang_2019_ICCV}                & 0.7768               & 0.7524          & 0.7230          & 0.6509          & 0.5705          & 0.7084           & 0.7593                 & 0.7308                \\
DGCNN\cite{dgcnn}+GCE\cite{zhang2018generalized}               & 0.7067               & 0.7003          & 0.6997          & 0.6967          & 0.6880          & 0.6614           & 0.7024                 & 0.7006                \\
DGCNN\cite{dgcnn}+SELFIE\cite{song2019selfie}*            & 0.8673               & 0.8158          & 0.7914          & 0.7725          & 0.7163          & 0.7500           & 0.8565                 & 0.8095                \\
DGCNN\cite{dgcnn}+PNAL               & 0.8686               & \textbf{0.8569} & \textbf{0.8378} & \textbf{0.8236} & \textbf{0.7651} & \textbf{0.7968}  & \textbf{0.8635}                 & \textbf{0.8498}                \\ \midrule
PointNet2\cite{qi2017pointnetplusplus}+CE             & \textbf{0.8898}      & 0.7008          & 0.6796          & 0.5850          & 0.5204          & 0.5648           & 0.8301                 & 0.7442                \\
PointNet2\cite{qi2017pointnetplusplus}+PNAL           & 0.8852               & \textbf{0.8385} & \textbf{0.8271} & \textbf{0.8067} & \textbf{0.7708} & \textbf{0.8202}  & \textbf{0.8670}                 & \textbf{0.8321}                \\ \bottomrule
\end{tabular}
} 
\end{table*}

Table.~\ref{Tab:s3dis} shows the performance of baselines and PNAL under different backbones, noisy rates and noise types. 
\ysq{Consistent with our observations on real-world noisy dataset ScanNetV2, we designed the ``\textit{mixed asymmetric noise}'' in Table.~\ref{Tab:s3dis} as a combination of both asymmetric (noise rate $\tau_{pair}=40\%$) and symmetric noise (noise rate $tau=60\%$). The presence of asymmetric noise is due to the similarity among the confusing categories. And symmetric noise exists because of the wrong category selection during the labeling process, i. e., as shown in \fref{teaser}, where the floor is labeled as a chair. Thus, the annotator introduces noise not only in the confusing classes, but also in the other classes, with random annotation errors.}

\ysq{Different from the ``\textit{mixed asymmetric noise}'' ($\tau_{pair}=40\%,\tau=60\%$) setting, which is a combination of asymmetric and symmetric noise, here we add more experiments on \textbf{pure} asymmetric noise following the similar setting in the previous image noisy label learning works.
In other words, we only randomly flip the label with a probability $\tau$ between our identified easily misclassified label pairs, including door-wall, board-window, and sofa-chair, while the labels of the other classes remain unchanged.}

The first five rows show the results with DGCNN backbone. 
In the case of DGCNN+CE, its performance drops quickly by 11.86\% at 20\% noise rate alone, \ysq{by 23\% at 60\% symmetric noise rate, by 30.58\% at mixed asymmetric noise, and by 11.79\% at $40\%$ pure asymmetric noise rate} compared to the result on clean training data.  This shows that label noise hurts the training process severely.
\ysq{The decline on pure asymmetric noise is not as dramatic as with mixed symmetric noise, because noises in this traditional asymmetric setting are only present in the easily confusing classes, which are a small percentage of the whole data.}

For the second and third row, we observe that previous noise-robust methods SCE and GCE perform no better or worse than CE at 0\% and 20\% noise rates, and only 1.19\% and 5.77\% improvement at 60\% symmetric noise rate. \ysq{They fail to work and even produce worse results on pure asymmetric noise. These are expected since these methods work with small or constant noise rates, while point cloud training suffers from extreme noise rate variation. And they do not consider the label correlation of local regions, so it is difficult to achieve optimal results.}

In the fourth row, the noise correction method SELFIE helps improve the performance over the CE results by 13.35\% at 60\% symmetric noise rate, by 18.66\% at mixed asymmetric noise, and by 5.82\% at 40\% pure asymmetric noise. \ysq{While these improvements are quite significant compared to SCE and GCE, the proposed PNAL can produce much better results even than SELFIE. We attribute this to the noise-blind design and local region correlation modeling.}

Compared to the DGCNN+SELFIE framework, DGCNN+PNAL shows a further improvement of more than 4.11\% for all noise settings.  It is worth noting that SELFIE requires a noise rate and takes more than 10 hours per epoch on average, while DGCNN+PNAL takes only 3 hours and 51 minutes. We owe this to the noise-blind pipeline, and the voting design that takes into account the label correlation of local regions. Furthermore, our method significantly improves result by 10.63\%, 16.46\%, 18.46\%, 25.91\%, 23.34\%, 9.85\% at 20\%, 40\%, 60\%, 80\% symmetric noise, mixed asymmetric noise and 40\% pure asymmetric noise, respectively.  
Consistently, our performance improves significantly by 13.77\%, 14.75\%, 22.17\%, 25.04\%, 25.54\%, and 8.79\% with PointNet2 as backbone, as shown in the last two rows.

\begin{figure*}[t] 
\centering
\setlength{\tabcolsep}{0.2mm}{
\begin{tabular}{ccccc}
\includegraphics[width=0.190\textwidth,page=1]{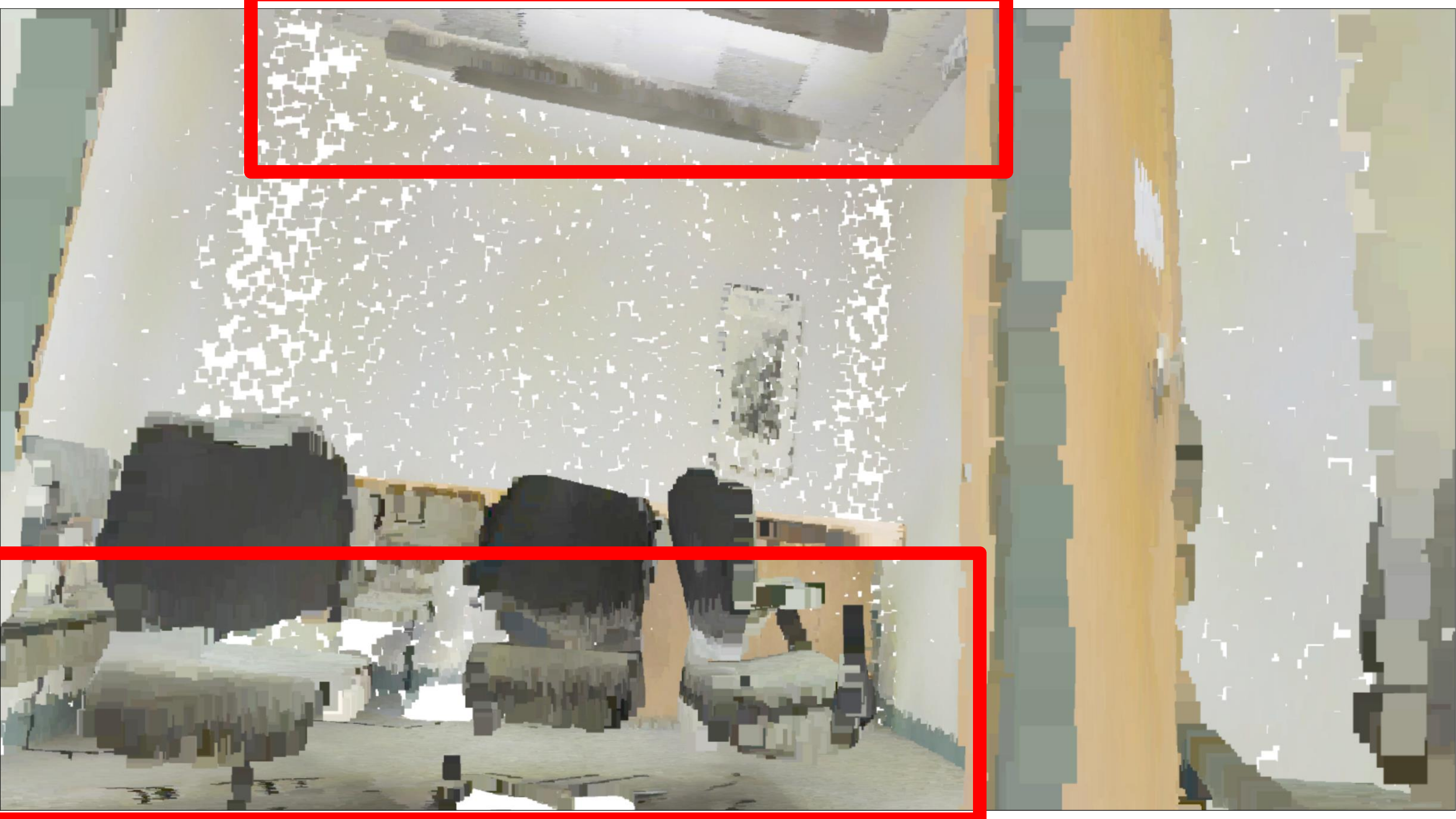} &
\includegraphics[width=0.190\textwidth,page=2]{LaTeX/img/s3dis_our_vs_dgcnn/Area_6_conferenceRoom_1.pdf} &
\includegraphics[width=0.190\textwidth,page=3]{LaTeX/img/s3dis_our_vs_dgcnn/Area_6_conferenceRoom_1.pdf} & 
\includegraphics[width=0.190\textwidth,page=4]{LaTeX/img/s3dis_our_vs_dgcnn/Area_6_conferenceRoom_1.pdf}  &  \multirow[c]{3}{*}[1.7cm]{
  \includegraphics[width=0.07\textwidth]{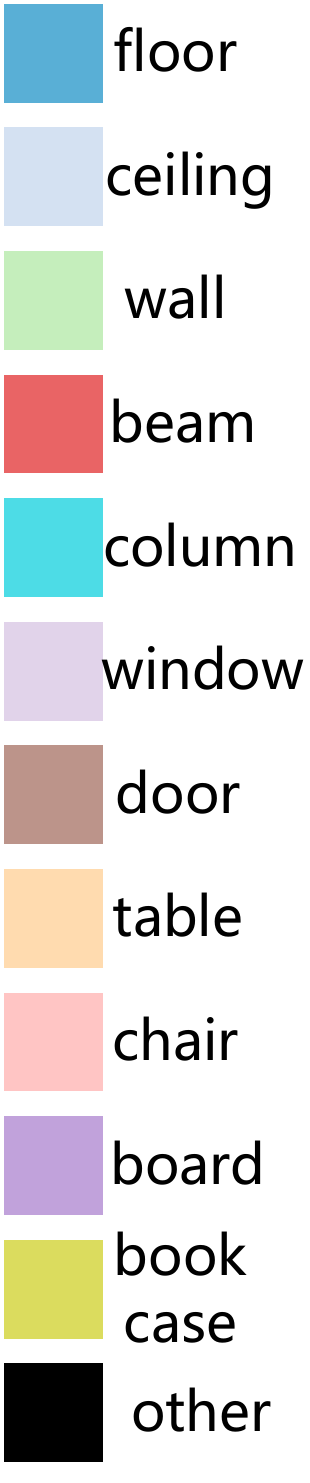}} \\ 
\includegraphics[width=0.190\textwidth,page=1]{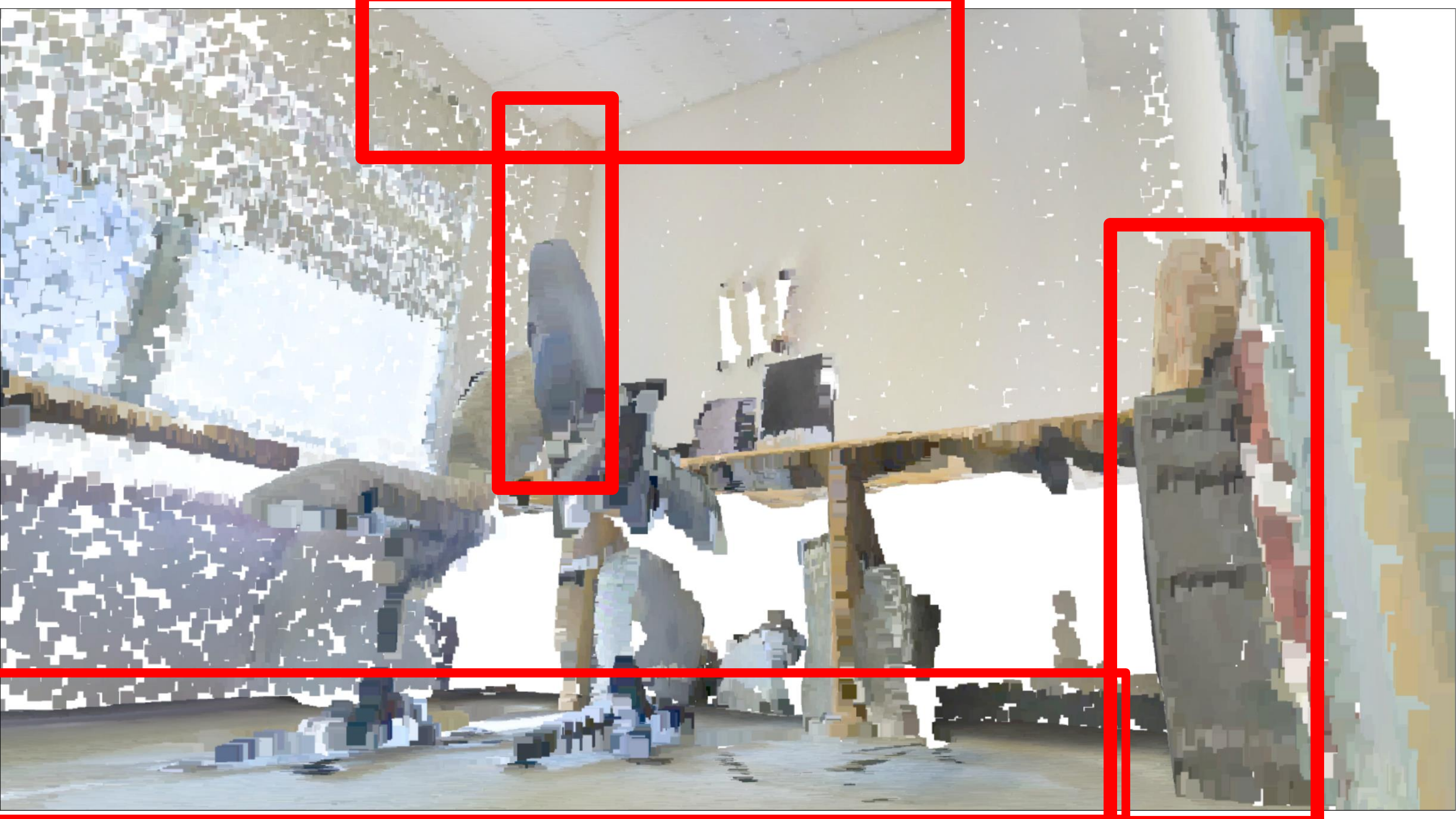} &
\includegraphics[width=0.190\textwidth,page=2]{LaTeX/img/s3dis_our_vs_dgcnn/Area_6_office_30.pdf} &
\includegraphics[width=0.190\textwidth,page=3]{LaTeX/img/s3dis_our_vs_dgcnn/Area_6_office_30.pdf} & 
\includegraphics[width=0.190\textwidth,page=4]{LaTeX/img/s3dis_our_vs_dgcnn/Area_6_office_30.pdf} 
& \\
\includegraphics[width=0.190\textwidth,page=1]{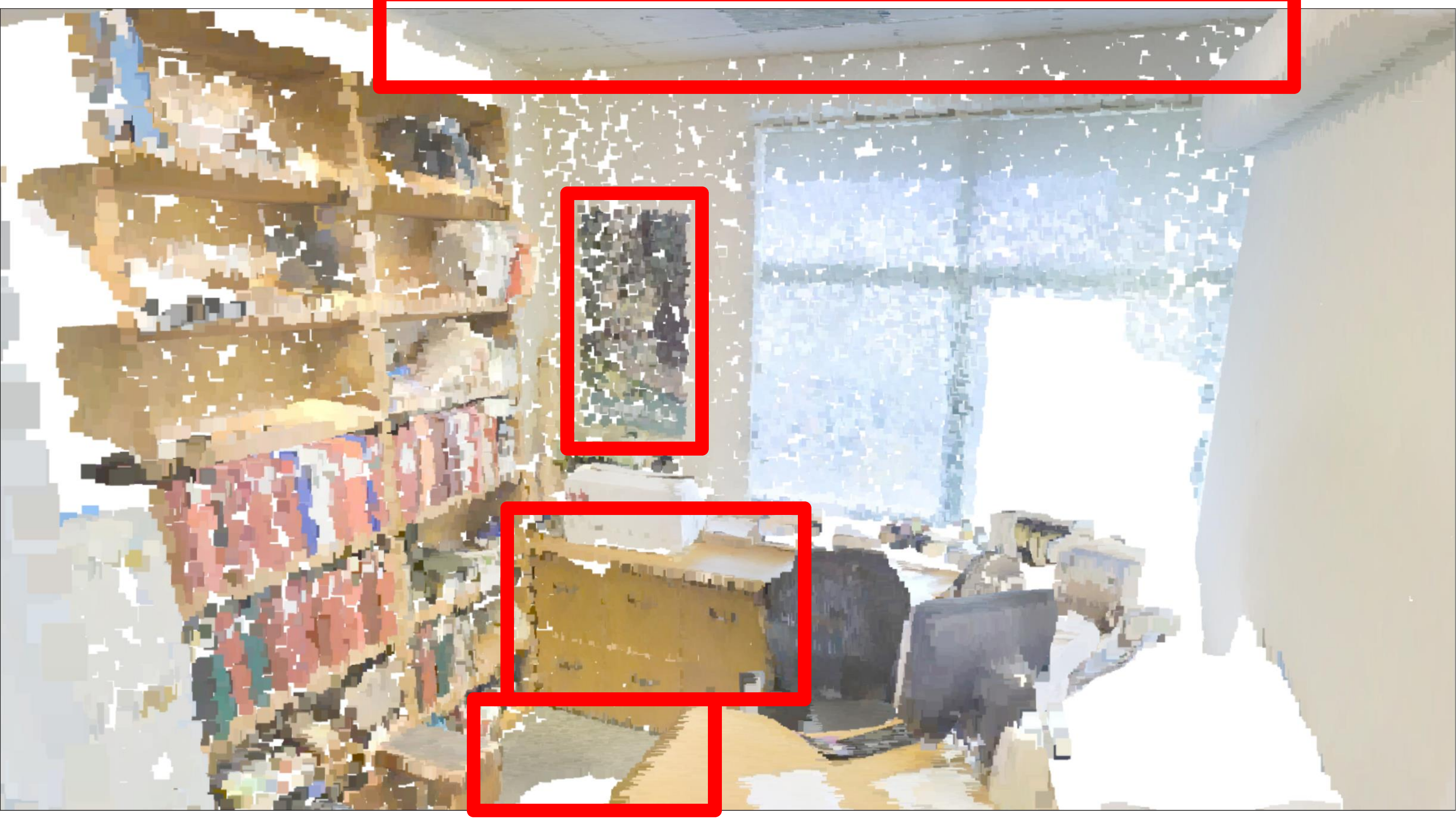} &
\includegraphics[width=0.190\textwidth,page=2]{LaTeX/img/s3dis_our_vs_dgcnn/Area_6_office_31.pdf} &
\includegraphics[width=0.190\textwidth,page=3]{LaTeX/img/s3dis_our_vs_dgcnn/Area_6_office_31.pdf} & 
\includegraphics[width=0.190\textwidth,page=4]{LaTeX/img/s3dis_our_vs_dgcnn/Area_6_office_31.pdf} 
& \\
 Input Scenes & GT Labels & DGCNN+CE & DGCNN+PNAL\\

\end{tabular}
}
\caption{From left to right: Scenes in S3DIS testset, clean GTs, predictions of DGCNN+CE, and DGCNN+PNAL.}
\label{Fig:s3distest}
\end{figure*}

\subsubsection{\ysq{Boundary-level Label Noise}}

\begin{table*}[ht]
\centering
\caption{OA Comparison on artificially created boundary noise S3DIS. The tops with different backbones are shown in bold.}
\setlength{\tabcolsep}{0.5mm}{
\scalebox{0.96}{
\begin{tabular}{@{}c|ccccccccc|ccccccccc@{}}
\toprule
\multirow{3}{*}{Methods}  & \multicolumn{9}{c|}{$\alpha = 0.5$}                                  & \multicolumn{9}{c}{$\alpha = 1.0$} \\ 
                 & \multicolumn{3}{c|}{$\beta = 0.3$}            & \multicolumn{3}{c|}{$\beta = 0.5$}           & \multicolumn{3}{c|}{$\beta = 0.7$}            & \multicolumn{3}{c|}{$\beta = 0.3$}            & \multicolumn{3}{c|}{$\beta = 0.5$} & \multicolumn{3}{c}{$\beta = 0.7$}            \\ 
                 \cmidrule(l){2-10}\cmidrule(l){11-19}
 & OA & @edge & \multicolumn{1}{c|}{@in} & OA & @edge & \multicolumn{1}{c|}{@in} & OA & @edge & \multicolumn{1}{c|}{@in} & OA & @edge & \multicolumn{1}{c|}{@in} & OA & @edge & \multicolumn{1}{c|}{@in} & OA & @edge & \multicolumn{1}{c}{@in} \\
\midrule 
DGCNN\cite{dgcnn}+CE      &  0.7702 & 0.5712 & 0.7944 & 0.7396 & 0.5498 & 0.7627 & 0.6614 & 0.4825 & 0.6831 & 0.6764 & 0.5071 & 0.6970 & 0.6646 & 0.4853 & 0.6864 & 0.6559 & 0.4848 & 0.6767    \\
DGCNN+SCE\cite{Wang_2019_ICCV}         &  0.7634 & 0.5696 & 0.7870 & 0.7298 & 0.5387 & 0.7531 & 0.7019 & 0.5175 & 0.7244 & 0.7021 & 0.5250 & 0.7237 & 0.7006 & 0.5149 & 0.7231 & 0.6704 & 0.4936 & 0.6918        \\
DGCNN+GCE\cite{zhang2018generalized}         &  0.7009 & 0.5010 & 0.7251 & 0.6992 & 0.5002 & 0.7233 & 0.6988 & 0.5121 & 0.7215 & 0.6984 & 0.5248 & 0.7196 & 0.6871 & 0.5022 & 0.7095 & 0.6626 & 0.4920 & 0.6833       \\
DGCNN+SELFIE\cite{song2019selfie}*  & 0.8223 & 0.6863 & 0.8394 & 0.7933 & 0.6205 & 0.8146 & 0.7575 & 0.5459 & 0.7831 & 0.7462 & 0.5310 & 0.7722 & 0.7150 & 0.5108 & 0.7397 & 0.6751 & 0.4910 & 0.6974       \\
DGCNN+PNAL        &  0.8426 & 0.6971 & 0.8608 & 0.8342 & 0.6630 & 0.8553 & 0.7981 & 0.5926 & 0.8256 & 0.8010 & 0.6315 & 0.8219 & 0.7857 & 0.5802 & 0.8107 & 0.6992 & 0.5053 & 0.7227        \\
DGCNN+PNAL-boundary          & \textbf{0.8662} & \textbf{0.7204} & \textbf{0.8845} & \textbf{0.8639} & \textbf{0.7203} & \textbf{0.8820} & \textbf{0.8620} & \textbf{0.7085} & \textbf{0.8812} & \textbf{0.8622} & \textbf{0.7161} & \textbf{0.8806} & \textbf{0.8600} & \textbf{0.7028} & \textbf{0.8795} & \textbf{0.8536} & \textbf{0.6993} & \textbf{0.8729} \\ 
\midrule
PointNet2\cite{qi2017pointnetplusplus}+CE               & 0.7332 & 0.5332 & 0.7575 & 0.7083 & 0.5176 & 0.7314 & 0.6675 & 0.4880 & 0.6893 & 0.6700 & 0.5072 & 0.6899 & 0.6694 & 0.4901 & 0.6912 & 0.6661 & 0.4896 & 0.6875      \\
PointNet2+PNAL-boundary        & \textbf{0.8802} & \textbf{0.7389} & \textbf{0.8980} & \textbf{0.8784} & \textbf{0.7448} & \textbf{0.8953} & \textbf{0.8726} & \textbf{0.7261} & \textbf{0.8910} & \textbf{0.8755} & \textbf{0.7295} & \textbf{0.8938} & \textbf{0.8740} & \textbf{0.7170} & \textbf{0.8937} & \textbf{0.8681} & \textbf{0.7166} & \textbf{0.8871} \\ 
\bottomrule
\end{tabular}
}
} 
\label{Tab:s3disboundary}
\end{table*}

\ysq{Table.~\ref{Tab:s3disboundary} shows the performance of baselines, PNAL and PNAL-boundary under boundary label noise with different dataset noise rates $\alpha$, sample noise rates $\beta$, and with different backbones. 
To make a more detailed comparison of the model performance in the boundary and inner areas separately, we further calculate the overall accuracy in the boundary and inner areas, i.e. A@edge and A@in. Based on the calculation of the overall accuracy of each test-set point cloud sample $\mathbf{X}$ as $OA(X) = \frac{|\mathbf{f}_{\theta}(\mathbf{X}) = \mathbf{Y}|} {|\mathbf{Y}|}$, we have the boundary and inner accuracy defined as: }

\begin{equation} \label{eq:8}
\begin{aligned}
&OA@edge = OA(\mathbf{X}_{boundary}),\\
&OA@in = OA(\mathbf{X}-\mathbf{X}_{boundary})\frysq{,}
\end{aligned}
\end{equation}
\ysq{, where $\mathbf{X}-\mathbf{X}_{boundary}$ is the point set of inner area.}

\ysq{The first six rows show the results with DGCNN backbone. 
In the case of DGCNN+CE, its performance drops quickly by 9.90\% at $0.5$ dataset noise and $0.3$ sample noise rate, and by 21.33\% at $1.0$ dataset noise and $0.7$ sample noise, compared to the result on clean training data.  This shows that the boundary label noise also hurts the training process severely.
We observe that previous noise-robust methods SCE and GCE perform no better or worse than CE at low noise rates ($\alpha=0.5$ and $\beta \leq 0.5$), and only 1.45\% and 0.67\% improvement at the highest noise rate of $\alpha=1.0$ and $\beta=0.7$. 
The SELFIE at the highest noise rate improves the performance by 1.92\%. 
We can draw similar conclusions as in the instance-level label noise that these methods work on small boundary noise, but cannot obtain optimal results on extreme noise rates. 
Compared to the DGCNN+SELFIE framework, the DGCNN+PNAL framework that proposed for instance label noise shows an improvement of about 3\% for all noise settings.  }

\ysq{After adopting the PNAL-boundary, we can observe a further improvement over PNAL framework of 5.3\% on average.
Respectively, our PNAL-boundary significantly improves the overall accuracy by 9.60\%, 12.43\%, 20.06\%, 18.58\%, 19.54\%, 19.77\% at each noise level. At the highest noise rate, the improvement on edge accuracy (from $0.4848$ to $0.6993$) even exceeds the improvement on inner accuracy (from $0.6767$ to $0.8729$) by 1.83\%, which demonstrates its robustness to boundary label noise.
Consistently, our performance improves by 14.70\%, 17.01\%, 20.51\%, 20.55\%, 20.46\%, 20.20\% with PointNet2 as backbone, as shown in the last two rows.}

\subsubsection{\ysq{Mixed Label Noise}}

\begin{table}[t] 
\caption{OA Comparison on artificially created noisy S3DIS of mixed instance and boundary noise with different levels. It demonstrates the robustness of the combination of the two frameworks on extreme mixed noisy setting.} 
\label{Tab:mixnoise}
\setlength{\tabcolsep}{0.75mm}{
\begin{tabular}{c|cc|cc}
\hline
                                                                                                      & DGCNN+CE & DGCNN+ours & PointNet2+CE & PointNet2+ours \\ \hline 
                                                                              \begin{tabular}[c]{@{}c@{}}$\tau$ =20\%,\\ $\alpha$=0.5, $\beta$=0.3\end{tabular} & 0.7372   & 0.8505     & 0.6698       & 0.8294         \\ \hline
 \begin{tabular}[c]{@{}c@{}}$\tau$ =40\%,\\ $\alpha$=1.0, $\beta$=0.5\end{tabular} & 0.6059   & 0.8230     & 0.5906       & 0.8033        \\ \hline
\begin{tabular}[c]{@{}c@{}}$\tau$ =80\%,\\ $\alpha$=1.0, $\beta$=0.7\end{tabular} & 0.4977   & 0.7137     & 0.5128       & 0.7501         \\ 

\hline
\end{tabular}} 
\end{table}

\ysq{To further investigate our framework quantitatively, we adopt a mixed label noise generation approach to more closely mimic real-world noise. We combined two noise models, i.e., creating instance-level label noise first and then boundary label noise, and generated the mixed noise manually on the S3DIS dataset. For each noise type, we use the noise rates of different levels, i.e., different instance-level symmetric noise rates $\tau$ and boundary noise rates $\alpha$ and $\beta$. To cope with the mixed noise, our method first applies the PNAL framework to train 30 epoches and then the PNAL-boundary framework for 10 epoches. 
The results are as shown in \tref{Tab:mixnoise}. Despite the extreme mixed noise conditions in the last row, i.e., 80\% instance-level symmetric noise and $\alpha=1.0, \beta=0.7$ boundary noise, our method consistently boosts the performance by more than 20\%. We also analyze the visualization of the label correction process on the mixed noise in \sref{subsec:corrprocess_mixed}.
}

\subsection{Experiments on Real-World Noisy ScanNetV2}

\subsubsection{Statistics of Noisy Label on ScanNetV2}

\begin{table}[t]
\centering
\caption{Statistics of instance-level noisy label on real-world noisy ScanNetV2 validation set. The label noise issue is widely present.}
\setlength{\tabcolsep}{4.4mm}{
\begin{tabular}{@{}c|c|c|c@{}} 
\toprule
Dataset   & \# Scenes & \# Noisy Labeled Scenes & Ratio \\ \midrule
ScanNetV2 & 312       & 228                     & $73.3\%$\\ \bottomrule
\end{tabular}}
\label{Tab:statistics}
\end{table}

\ysq{For a better understanding of the instance-level label noise issue on the ScanNetv2 dataset, we leverage manual inspection to check the whole ScanNetV2 validation set. For each scene, there were at least two human inspectors and the scene is counted as mislabeled when they both agreed that \textbf{instance-level} noisy class labels exist in the scene. As shown in Table \ref{Tab:statistics}, 228 scenes out of 312 (73.3\%) suffer from instance-level label noise issue. Therefore, even though ScanNetV2 is already a re-labeled version of ScanNet,  the label noise issue is widely present in its validation set. This further proves the existence and significance of the noisy labeling problem and the urgency of the need for a noise-robust framework.}

\subsubsection{Performance of ScanNetV2}

\begin{figure*}[!h] 
\centering
\setlength{\tabcolsep}{1.25mm}{
\scalebox{1.1}{ 
\begin{tabular}{ccccc}
\vspace{-1.55mm} 
\includegraphics[width=0.190\textwidth,page=1]{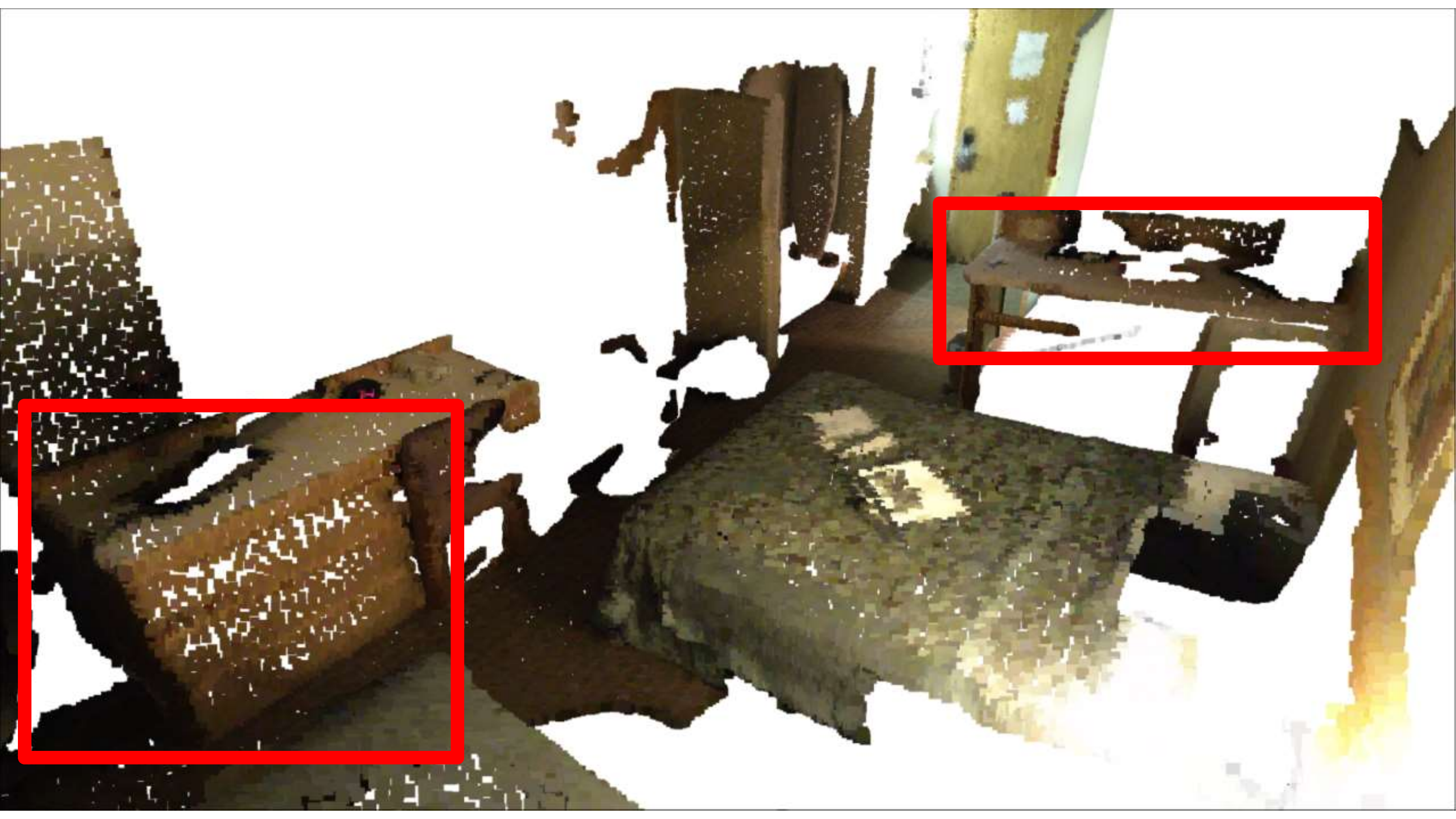} &
\includegraphics[width=0.190\textwidth,page=2]{LaTeX/img/scannetv2_our_vs_gt_vs_scn/small_size_72_oa0.8834_oa20.9227.pdf} &
\includegraphics[width=0.190\textwidth,page=3]{LaTeX/img/scannetv2_our_vs_gt_vs_scn/small_size_72_oa0.8834_oa20.9227.pdf} & 
\includegraphics[width=0.190\textwidth,page=4]{LaTeX/img/scannetv2_our_vs_gt_vs_scn/small_size_72_oa0.8834_oa20.9227.pdf} & 
\vspace{-1.55mm} 
\\ 
\includegraphics[width=0.190\textwidth,page=1]{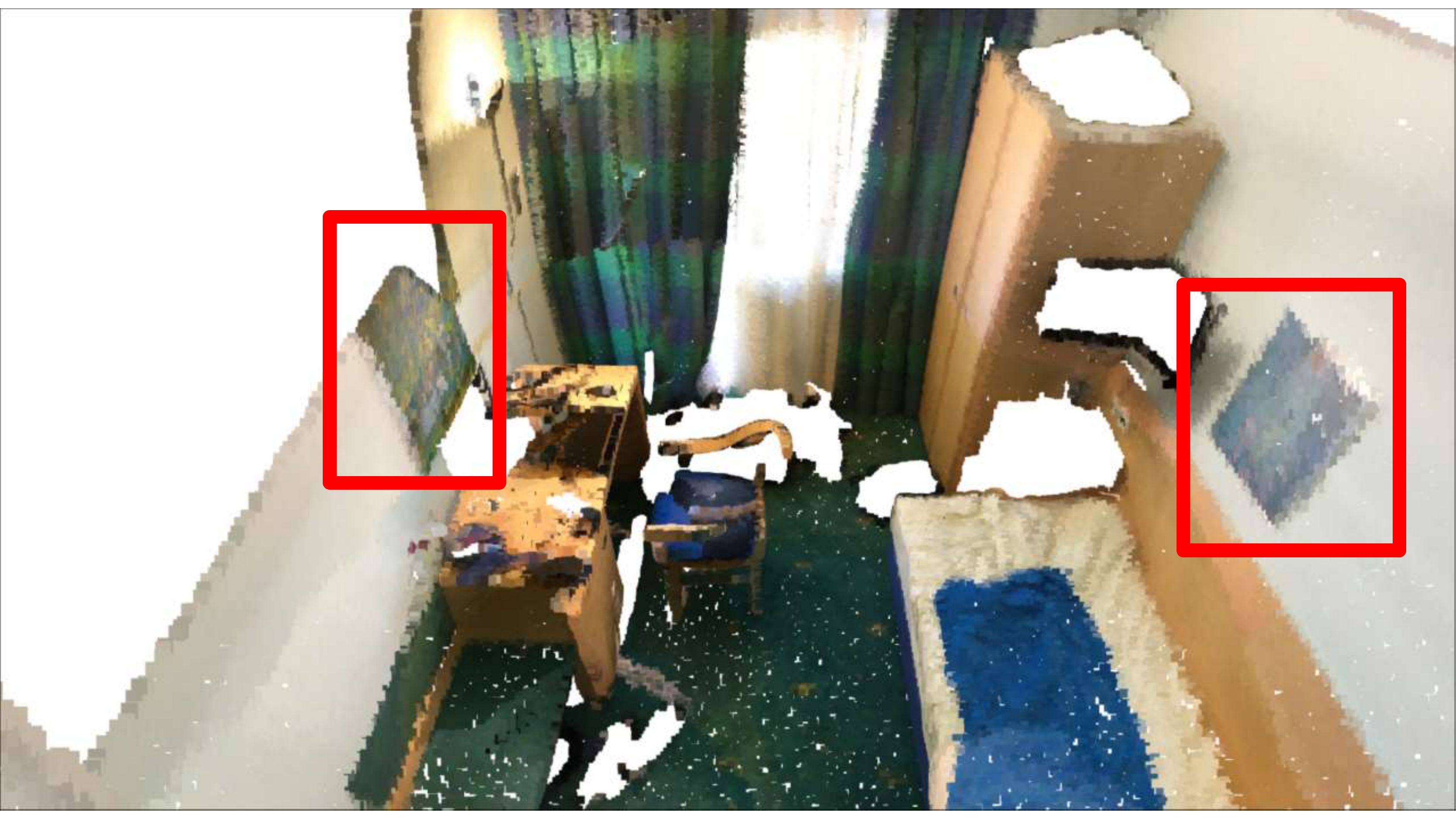} &
\includegraphics[width=0.190\textwidth,page=2]{LaTeX/img/scannetv2_our_vs_gt_vs_scn/small_size_13_oa0.8621_oa20.9072.pdf} &
\includegraphics[width=0.190\textwidth,page=3]{LaTeX/img/scannetv2_our_vs_gt_vs_scn/small_size_13_oa0.8621_oa20.9072.pdf} & 
\includegraphics[width=0.190\textwidth,page=4]{LaTeX/img/scannetv2_our_vs_gt_vs_scn/small_size_13_oa0.8621_oa20.9072.pdf} 
\vspace{-1.55mm} 
& \\
\includegraphics[width=0.190\textwidth,page=1]{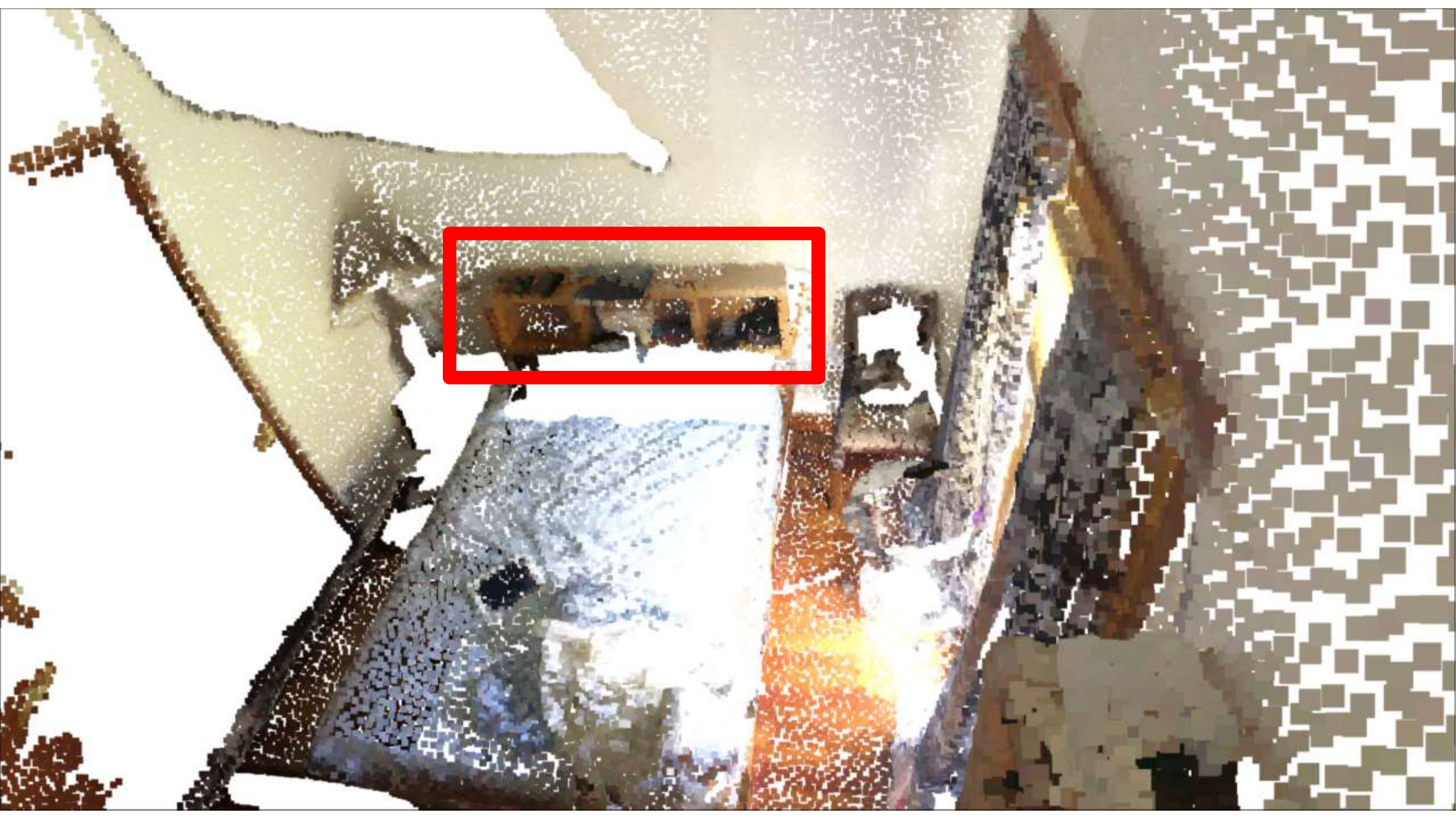} &
\includegraphics[width=0.190\textwidth,page=2]{LaTeX/img/scannetv2_our_vs_gt_vs_scn/small_size_203_oa0.7819_oa20.8374.pdf} &
\includegraphics[width=0.190\textwidth,page=3]{LaTeX/img/scannetv2_our_vs_gt_vs_scn/small_size_203_oa0.7819_oa20.8374.pdf} & 
\includegraphics[width=0.190\textwidth,page=4]{LaTeX/img/scannetv2_our_vs_gt_vs_scn/small_size_203_oa0.7819_oa20.8374.pdf} &
\vspace{-1.55mm} 
\multirow[c]{ 9}{*}[1.8cm]{\includegraphics[width=0.102\textwidth]{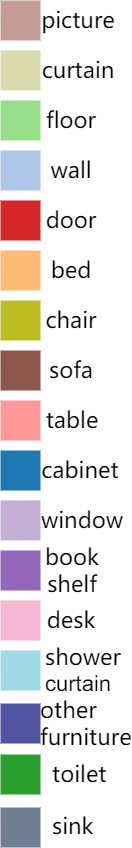}} 
\\

\includegraphics[width=0.190\textwidth,page=1]{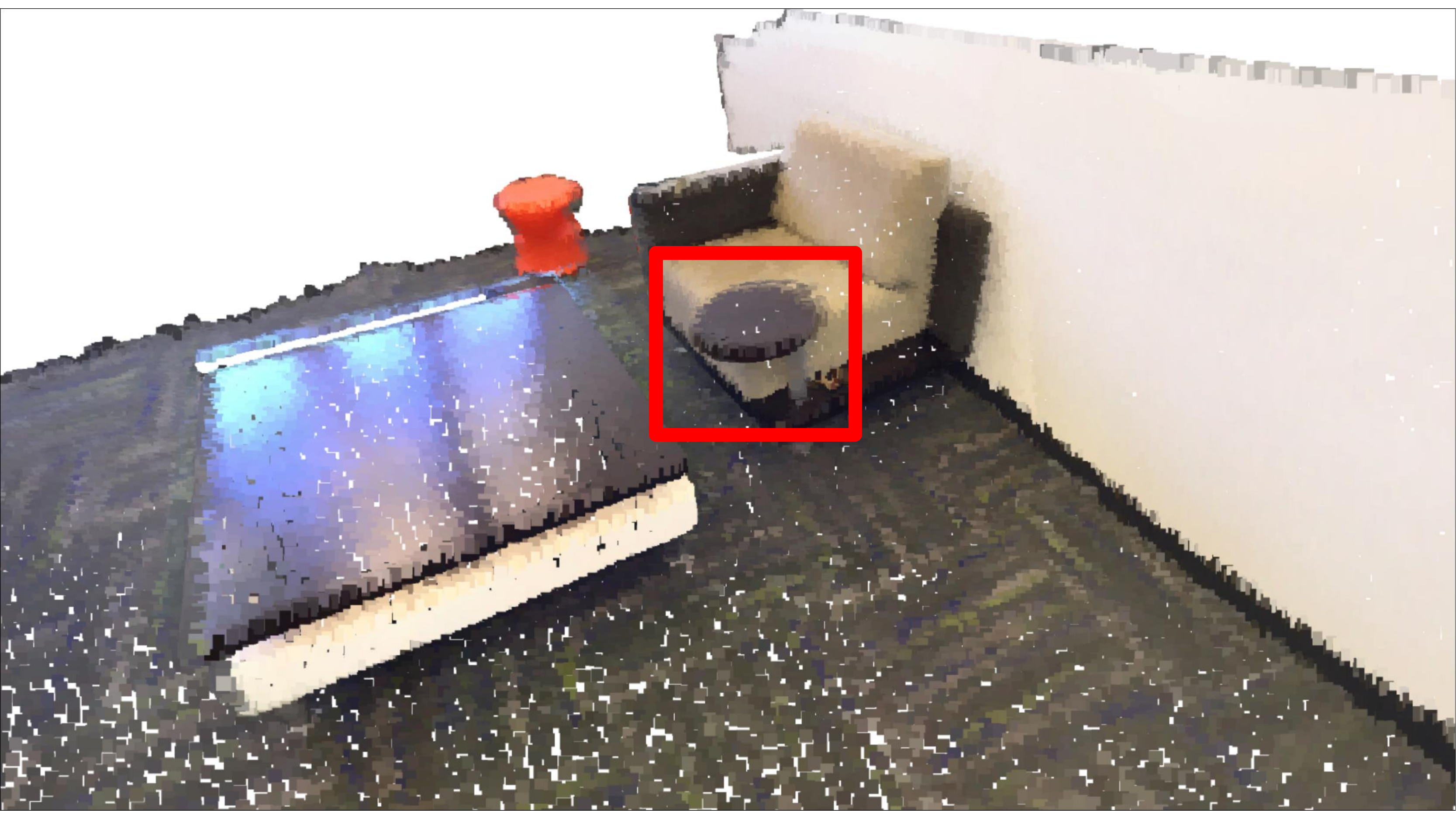} &
\includegraphics[width=0.190\textwidth,page=2]{LaTeX/img/supp_scannetv2_our_vs_gt_vs_scn/59_oa0.9506_oa20.7511.pdf} &
\includegraphics[width=0.190\textwidth,page=3]{LaTeX/img/supp_scannetv2_our_vs_gt_vs_scn/59_oa0.9506_oa20.7511.pdf} & 
\includegraphics[width=0.190\textwidth,page=4]{LaTeX/img/supp_scannetv2_our_vs_gt_vs_scn/59_oa0.9506_oa20.7511.pdf} & 
\vspace{-1.55mm} 
\\ 
\includegraphics[width=0.190\textwidth,page=1]{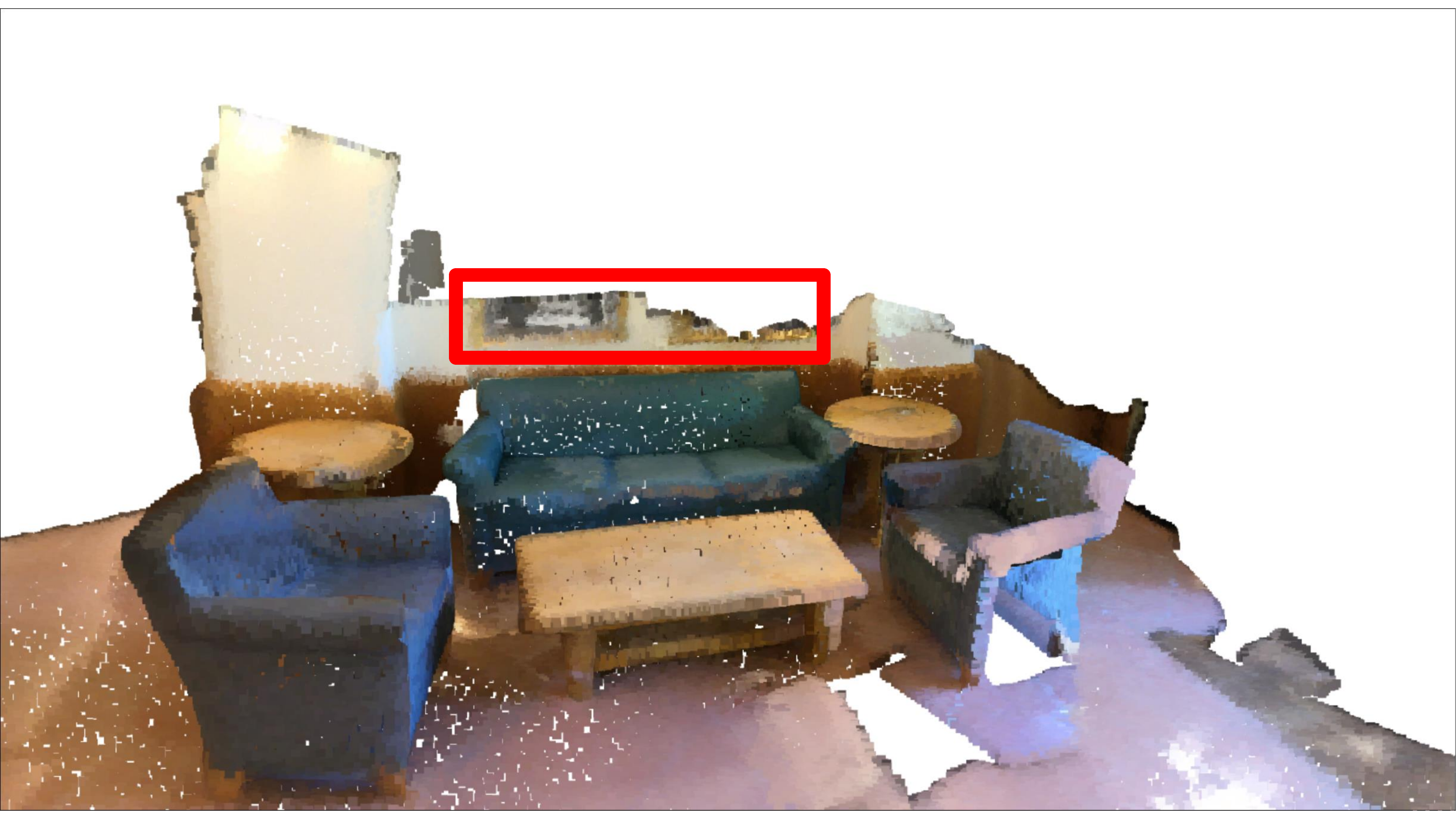} &
\includegraphics[width=0.190\textwidth,page=2]{LaTeX/img/supp_scannetv2_our_vs_gt_vs_scn/248_oa0.9654905796051025.pdf} &
\includegraphics[width=0.190\textwidth,page=3]{LaTeX/img/supp_scannetv2_our_vs_gt_vs_scn/248_oa0.9654905796051025.pdf} & 
\includegraphics[width=0.190\textwidth,page=4]{LaTeX/img/supp_scannetv2_our_vs_gt_vs_scn/248_oa0.9654905796051025.pdf} 
\vspace{-1.55mm} 
& \\
\includegraphics[width=0.190\textwidth,page=1]{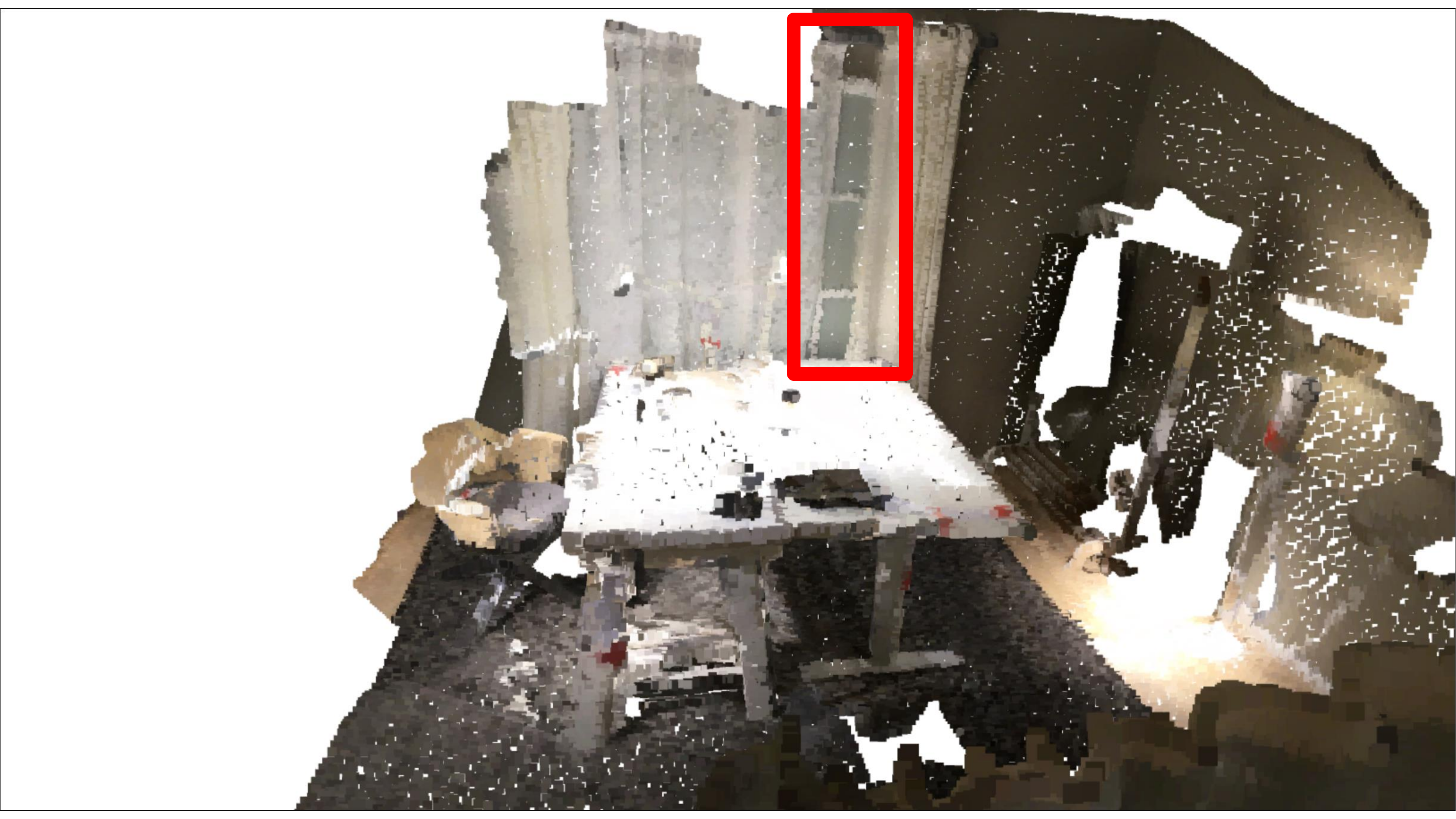} &
\includegraphics[width=0.190\textwidth,page=2]{LaTeX/img/supp_scannetv2_our_vs_gt_vs_scn/288_oa0.9463763236999512.pdf} &
\includegraphics[width=0.190\textwidth,page=3]{LaTeX/img/supp_scannetv2_our_vs_gt_vs_scn/288_oa0.9463763236999512.pdf} & 
\includegraphics[width=0.190\textwidth,page=4]{LaTeX/img/supp_scannetv2_our_vs_gt_vs_scn/288_oa0.9463763236999512.pdf} 
\vspace{-1.55mm} 
& \\
\includegraphics[width=0.190\textwidth,page=1]{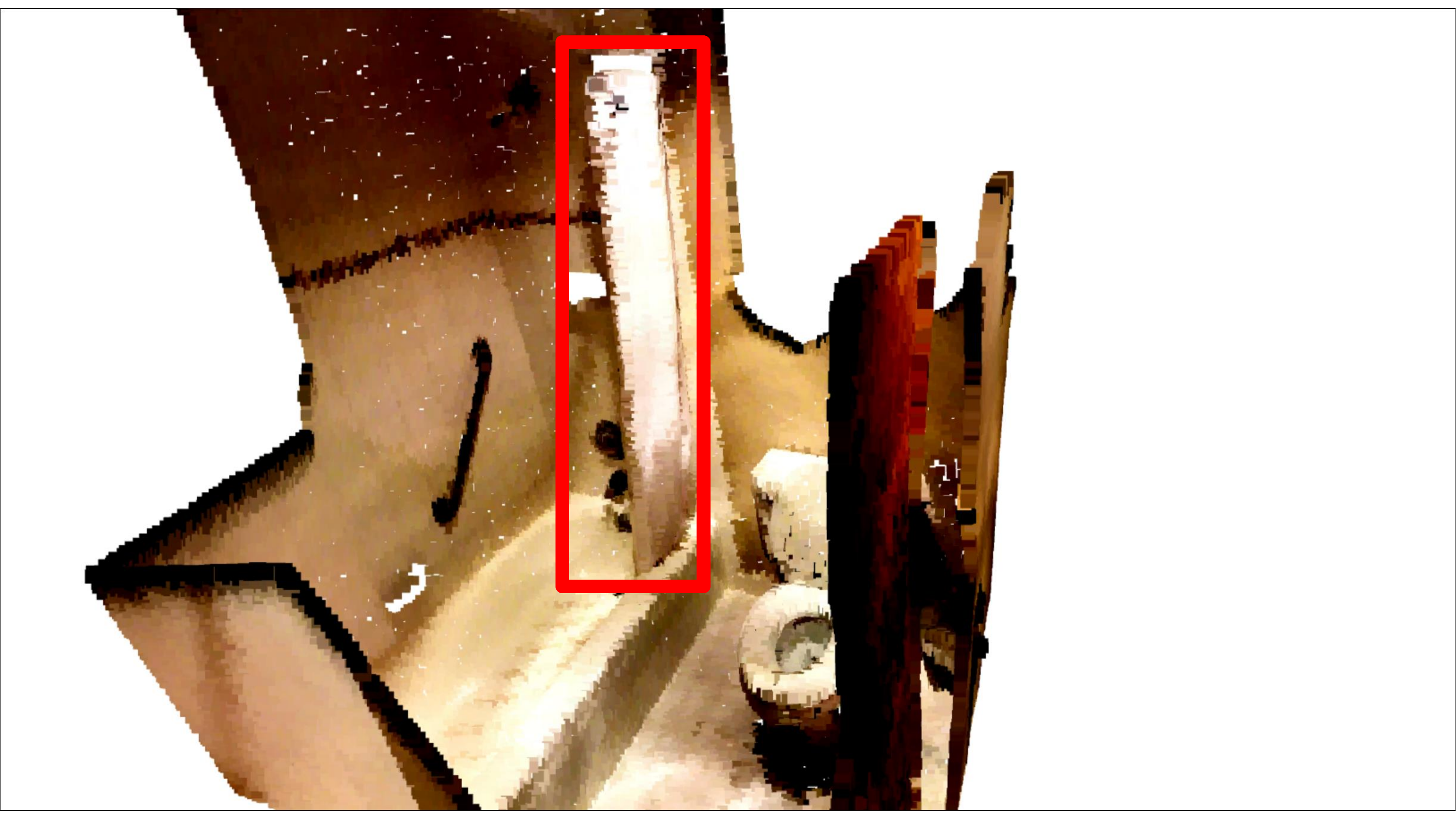} &
\includegraphics[width=0.190\textwidth,page=2]{LaTeX/img/supp_scannetv2_our_vs_gt_vs_scn/309_from_relabeled.pdf} &
\includegraphics[width=0.190\textwidth,page=3]{LaTeX/img/supp_scannetv2_our_vs_gt_vs_scn/309_from_relabeled.pdf} & 
\includegraphics[width=0.190\textwidth,page=4]{LaTeX/img/supp_scannetv2_our_vs_gt_vs_scn/309_from_relabeled.pdf} 
\vspace{-1.55mm} 
& \\

\includegraphics[width=0.1950\textwidth,page=1]{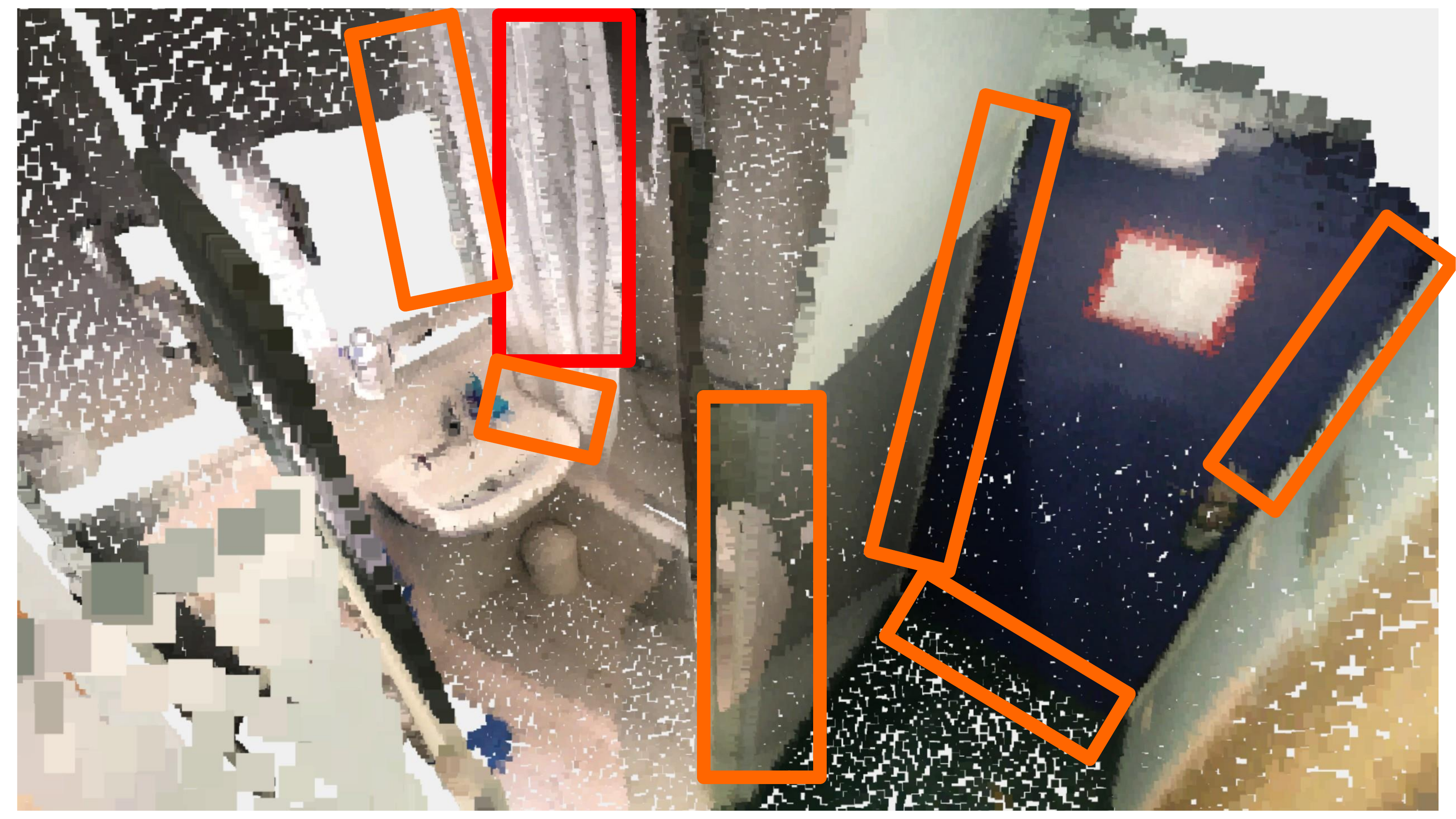} &
\includegraphics[width=0.1950\textwidth,page=2]{LaTeX/img/scannetv2_our_vs_gt_vs_scn/203_boundary.pdf} &
\includegraphics[width=0.1950\textwidth,page=3]{LaTeX/img/scannetv2_our_vs_gt_vs_scn/203_boundary.pdf} & 
\includegraphics[width=0.1950\textwidth,page=4]{LaTeX/img/scannetv2_our_vs_gt_vs_scn/203_boundary.pdf} & 
\vspace{-1.55mm} 
\\ 
\includegraphics[width=0.1950\textwidth,page=1]{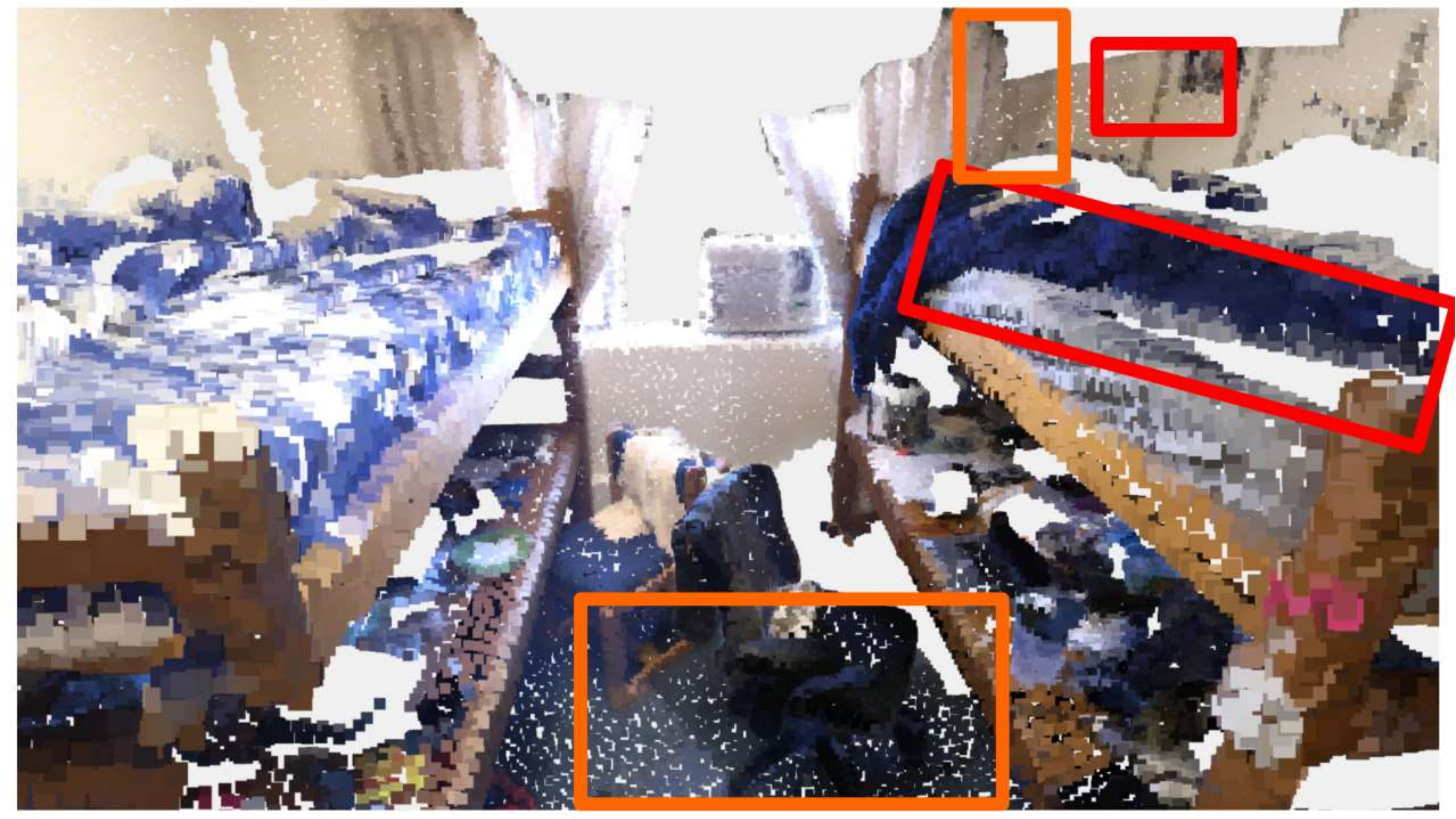} &
\includegraphics[width=0.1950\textwidth,page=2]{LaTeX/img/scannetv2_our_vs_gt_vs_scn/353_boundary.pdf} &
\includegraphics[width=0.1950\textwidth,page=3]{LaTeX/img/scannetv2_our_vs_gt_vs_scn/353_boundary.pdf} & 
\includegraphics[width=0.1950\textwidth,page=4]{LaTeX/img/scannetv2_our_vs_gt_vs_scn/353_boundary.pdf} & 
\vspace{-1.55mm} 
\\
\includegraphics[width=0.1950\textwidth,page=1]{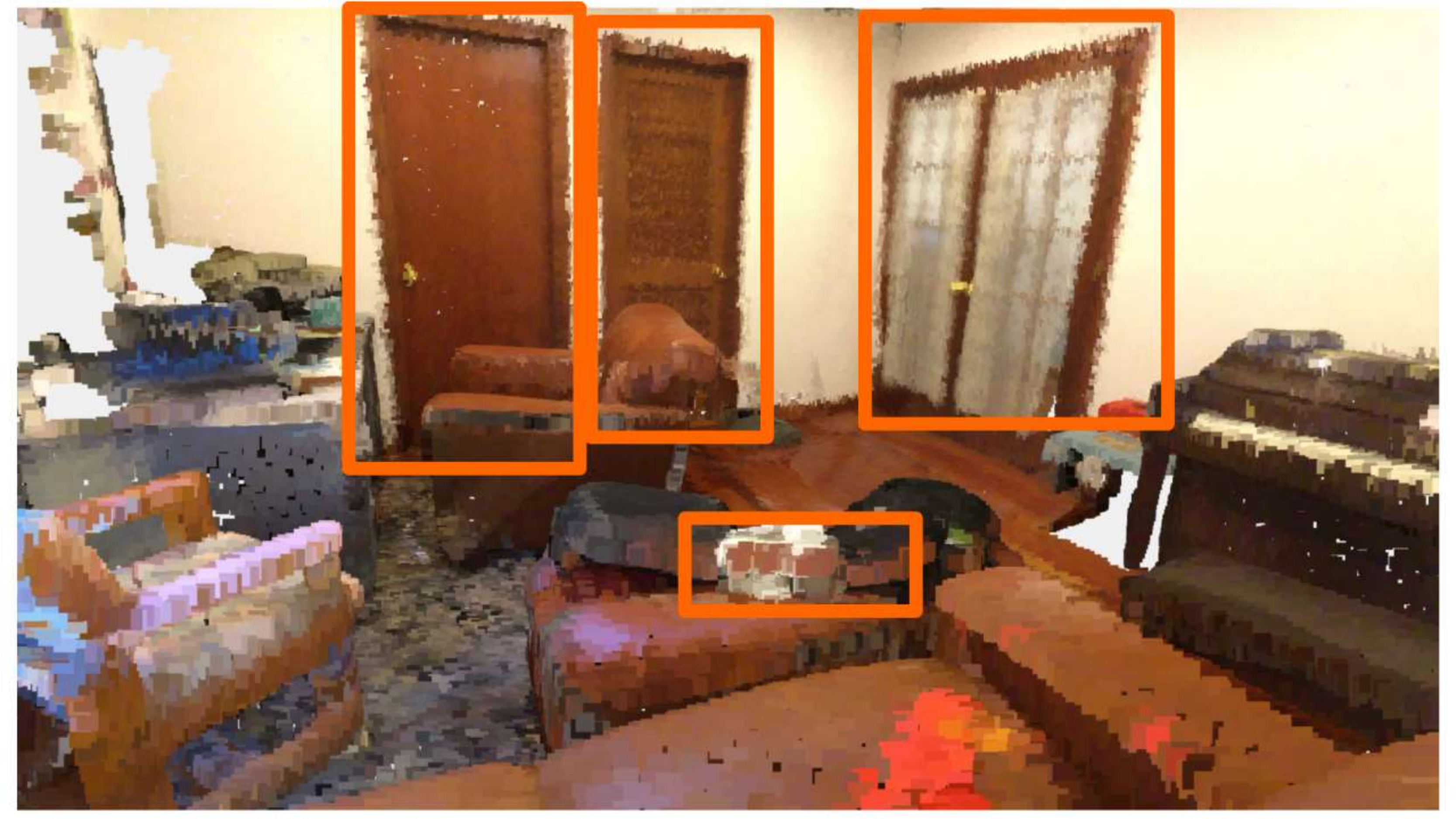} &
\includegraphics[width=0.1950\textwidth,page=2]{LaTeX/img/scannetv2_our_vs_gt_vs_scn/257_boundary.pdf} &
\includegraphics[width=0.1950\textwidth,page=3]{LaTeX/img/scannetv2_our_vs_gt_vs_scn/257_boundary.pdf} & 
\includegraphics[width=0.1950\textwidth,page=4]{LaTeX/img/scannetv2_our_vs_gt_vs_scn/257_boundary.pdf} & 
\vspace{-1.55mm} 
\\
\includegraphics[width=0.1950\textwidth,page=1]{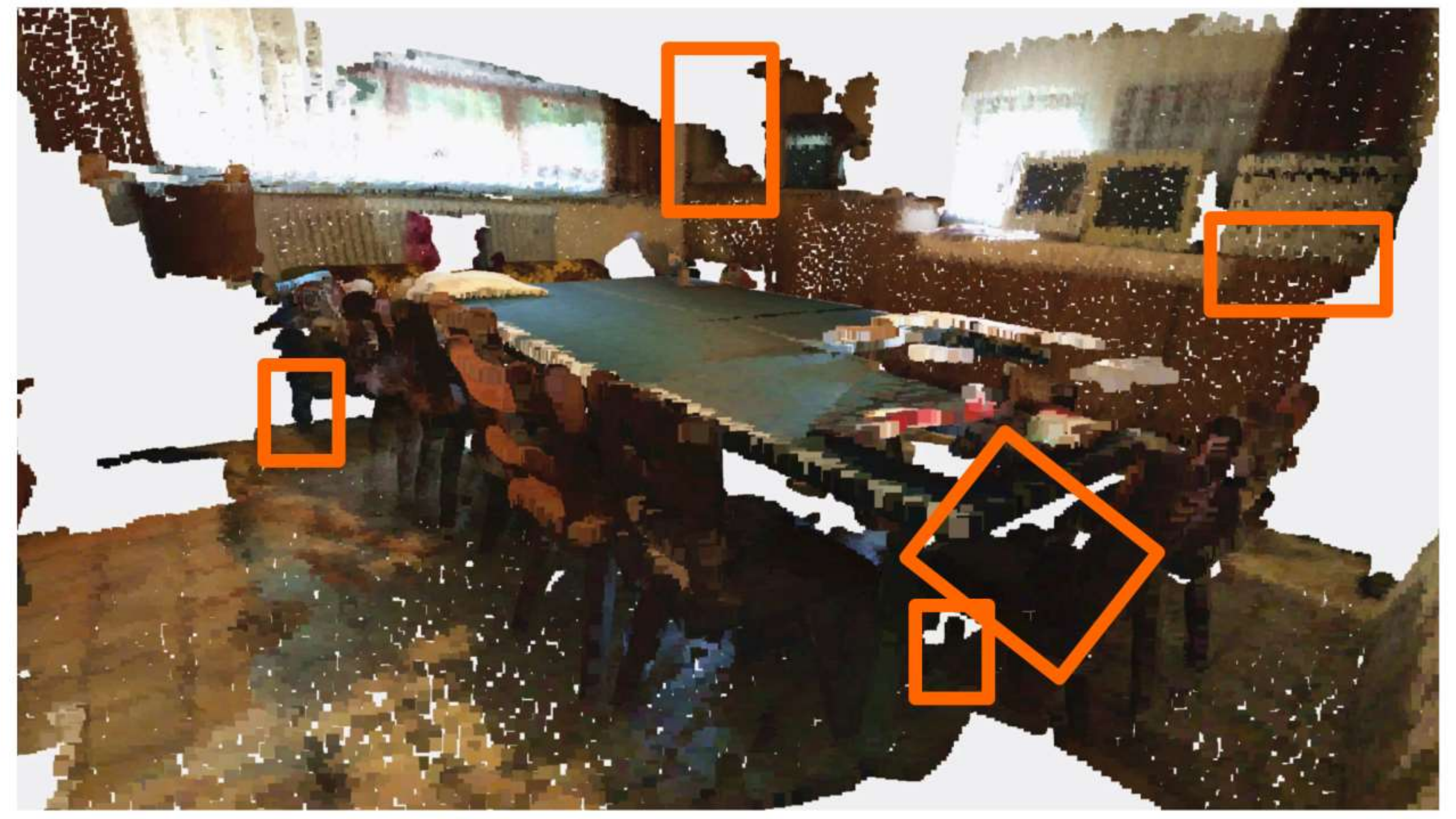} &
\includegraphics[width=0.1950\textwidth,page=2]{LaTeX/img/scannetv2_our_vs_gt_vs_scn/351_boundary.pdf} &
\includegraphics[width=0.1950\textwidth,page=3]{LaTeX/img/scannetv2_our_vs_gt_vs_scn/351_boundary.pdf} & 
\includegraphics[width=0.1950\textwidth,page=4]{LaTeX/img/scannetv2_our_vs_gt_vs_scn/351_boundary.pdf} & 
\vspace{-1.55mm} 
\\
 Input Scenes & GT Labels & SparseConvNet~\cite{3DSemanticSegmentationWithSubmanifoldSparseConvNet} & SparseConvNet~\cite{3DSemanticSegmentationWithSubmanifoldSparseConvNet}+ours\\

\end{tabular}
}}
\vspace{-1.5mm}
\caption{From left to right: Scenes in ScanNetV2 validation set, GT labels given by ScanNetV2, predictions of \cite{3DSemanticSegmentationWithSubmanifoldSparseConvNet}, and predictions of \cite{3DSemanticSegmentationWithSubmanifoldSparseConvNet}+ours. Ours get more reasonable labels than GT labels with both instance-level (in red boxes) and boundary-level (in orange boxes) label noise.}
\label{Fig:scannetv2test}
\end{figure*}

\ysq{We report the results of PNAL and PNAL-boundary framework on ScanNetV2 validation set with SparseConvNet \cite{3DSemanticSegmentationWithSubmanifoldSparseConvNet} as the backbone network. We note that the ScanNetV2 has both instance-level and boundary-level label noise on both training and validation splits.}

Table \ref{Tab:scannetv2} shows the performances of PNAL framework on ScanNetV2 validation set with SparseConvNet as the backbone network.
Although the results of our method do not show a great advantage, we believe this is due to the fact that there are still mislabels in the validation set of ScanNetV2. 
Therefore, we show some results of PNAL framework from the validation set in the first 7 rows of Fig. \ref{Fig:scannetv2test}, where PNAL gets more reasonable results than the baseline method and even than GT.

\begin{table}[t]
\caption{The mIoU and OA comparison on real-world noisy ScanNetV2 validation set and our re-labeled ScanNetV2 validation set.}
\centering
\label{Tab:scannetv2}
\scalebox{1.0}{
\begin{tabular}{@{}c|cccc@{}}
\toprule
\multicolumn{1}{c|}{\multirow{2}{*}{Methods}} & \multicolumn{2}{l}{\begin{tabular}[c]{@{}c@{}}real-world noisy \\ ScanNetV2\end{tabular}} & \multicolumn{2}{l}{\begin{tabular}[c]{@{}c@{}}our re-labeled\\ ScanNetV2\end{tabular}} \\ \cmidrule(l){2-5} 
\multicolumn{1}{c|}{}                         & mIoU                                        & OA                                          & mIoU                                       & OA                                        \\ \cmidrule(r){1-5}
SparseConvNet\cite{3DSemanticSegmentationWithSubmanifoldSparseConvNet}                                     & 0.7250                                      & 0.8928                                      & 0.7103                                     & 0.8807                                    \\
SparseConvNet\cite{3DSemanticSegmentationWithSubmanifoldSparseConvNet}+PNAL                                & 0.7298                                      & 0.8979                                      & 0.7416                                     & 0.9211                                    \\ \bottomrule
\end{tabular}
}

\end{table}

For a more rigorous comparison, we further tested on the fully relabeled clean validation data mentioned before, and reported the results in Table \ref{Tab:scannetv2}. Our method achieves significant performance gain, which demonstrates the effectiveness of our method on the real-world noisy dataset. In contrast, the performance of the baseline SparseConvNet degrades, indicating its overfitting of the label noise, which affects the performance on fully clean data.

\ysq{As ScanNetV2 validation set also suffers from boundary noise, while high quality data with accurate bundaries are time consuming to re-annotate, instead of quantitative experiments, we show some example results from the validation set. Note that, the fully relabeled clean validation data mentioned before focuses on instance-level label noise, and we keep boundary-level noisy labels untouched.
Our method first applies the PNAL framework to train 30 epoches and then the PNAL-boundary framework for 10 epoches, and the baseline is trained for 40 epoches.
The results are as shown in the last 4 rows of Fig. \ref{Fig:scannetv2test}, where ours gets more reasonable results than the baseline method and even than GT. Specifically, for the 8-th and 9-th rows, both instance-level and boundary-level label noise exists in the given labels, and our pipelines correct both of them properly.
These visualization results further illustrate the existence and complexity of real-world label noises. Specifically, the GT labeling error in the second and fourth rows are from confusing category pairs ``photos \textit{vs} walls''  and ``shower curtains \textit{vs} curtains'',which fit the asymmetric noise pattern. For the first and third row, unlike the aforementioned noise types, their labeling error pattern are sample-related, called instance- and label-dependent noises. Owing to its complexity,  modeling this kind of noise has not been extensively investigated yet. In the future, studying more diverse label noises in point cloud segmentation would be a great research direction.}

\subsection{Ablation Study}

\subsubsection{Ablation on PNAL}

\noindent\textbf{Component Ablation Study.}
\begin{table}[!t]
\centering
\caption{OA Comparison on $60\%$ symmetric noise S3DIS dataset.} 
\label{Tab:ablation}
\begin{tabular}{@{}c|cccccc@{}}
\toprule
Metric & \begin{tabular}[c]{@{}c@{}}GT \\ Instance\end{tabular} & \begin{tabular}[c]{@{}c@{}}w/o \\ Voting\end{tabular} & $\gamma$=1 & $\gamma$=2 & $q$=8  & \begin{tabular}[c]{@{}c@{}}DGCNN\\ +PNAL\end{tabular} \\ \midrule
OA      & \textbf{0.8287}                                        & 0.8011                                                & 0.8110     & 0.8209     & 0.7704 & 0.8236                                                \\ \bottomrule
\end{tabular}
\end{table}
All the results in Table. \ref{Tab:ablation} are on $60\%$ symmetric noise.
The first column reports the results of PNAL with GT instance instead of clustering for label correction, which represents the upper limit of our results. Compared with it, the cluster based results in the last column have only a small drop, which can illustrate the feasibility of using cluster as an alternative to GT instance label.
In the second column, we omit the cluster-level voting step and perform label correction point-wisely, without considering the label consistency between nearby points. The results show a $2.25\%$ decrease over our full method, demonstrating the effectiveness of our proposed cluster-level label correction.
In the third and fourth columns, we try different values of $\gamma$, where $\gamma=1$ is the most greedy case, i.e., the winner label is the top reliable label. We observe no significant performance drop with different $\gamma$ values, implying our method is not sensitive to the choice of $\gamma$. We use $\gamma =4$ in our setting. 
In the fifth column we adjust the history length $q$ to 8, and note that $E_{warm-up}$ is also increased to 8, due to the constraint of $q$. We can observe a significant decrease in performance. More analysis is given in the next paragraph.

\begin{table}[t]
\centering
\caption{The OA of PNAL at different $E_{warm-up}$ and noise rates.} \label{Tab:Ewarmupvsnoiserate}
\setlength{\tabcolsep}{7.0mm}
{
\begin{tabular}{@{}c|ccc@{}}
\toprule
Noise Rate $\tau$ & 20\%   & 40\%   & 60\%   \\ \midrule
$E_{warm-up}=5$         & \textbf{0.8569} & \textbf{0.8378} & \textbf{0.8236} \\
$E_{warm-up}=8$         & 0.8422 & 0.8247 & 0.7851 \\
$E_{warm-up}=11$        & 0.8343 & 0.8009 & 0.7812 \\ \bottomrule
\end{tabular} 
}
\end{table}

\begin{table}[t]
\centering
\caption{The OA of our method at different granularity of clusters and different clustering methods.}
\vspace{-1.2mm} 
\label{tab:granularity}
\setlength{\tabcolsep}{2.5mm}{
\begin{tabular}{@{}c|ccccc@{}}
\toprule
\begin{tabular}[c]{@{}c@{}}Clustering \\ Methods\end{tabular} & \begin{tabular}[c]{@{}c@{}}DBSCAN \\ $\varepsilon$=0.015\end{tabular} & \begin{tabular}[c]{@{}c@{}}DBSCAN \\ $\varepsilon$=0.018\end{tabular} & \begin{tabular}[c]{@{}c@{}}DBSCAN\\ $\varepsilon$=0.021\end{tabular} & GMM & spectral    \\ \midrule
OA                                                            & 0.8206                                                      & \textbf{0.8236}                                             & 0.8159                                                     & 0.8178 & 0.8162 \\ \bottomrule
\end{tabular}
}
\end{table}

\begin{table}[!t]
\centering
\vspace{-1.2mm} 
\caption{The OA of our method in an iterative manner.}
\vspace{-1.2mm} 
\label{tab:iterative}
\setlength{\tabcolsep}{5.8mm}{
\begin{tabular}{@{}c|ccc@{}}
\toprule
Iteration Start at epochs & 7,14,21 & 10,20 & 15    \\ \midrule
OA                                                            &  0.8054                                                     &   0.8115                                           &   0.8263                               \\ \bottomrule
\end{tabular}
}
\vspace{-1.5mm} 
\end{table}

\noindent\textbf{Robustness to $E_{warm-up}$.}\label{subsec:abl1_Ewarmupvsnoiserate}
Table \ref{Tab:Ewarmupvsnoiserate} reports the results of our method at different $E_{warm-up}$ under symmetric noise of different noise rates ($20\%, 40\%, 60\%$). Our performances are optimal and robust, with $E_{warm-up}=5$ as we recommend, for all noise rates, demonstrating that $E_{warm-up}$ is not sensitive to noise rate variations.
The larger the noise rate, the larger the performance drop can be observed if $E_{warm-up}$ increases. We can conclude that the larger the $E_{warm-up}$, the more noisy data the network fits, which can make noise-cleaning difficult on data with large noise rates.
Comparing the result in the second row with the fifth column in Tab.~\ref{Tab:ablation}, we found that although increasing the history length brings a performance decrease, this mainly comes from the effect of the increased $E_{warm-up}$.

\begin{table}[t]
\centering
\caption{OA Comparison on $\alpha=1.0, \beta=0.7$ boundary label noise S3DIS.}
\label{tab:ablation_boundary}
\setlength{\tabcolsep}{5.5mm}{
\begin{tabular}{@{}c|cccc@{}}
\toprule
Metric & \begin{tabular}[c]{@{}c@{}}w/o\\ Progressive\end{tabular} & k=10   & k=20   & k=30   \\ \midrule
OA     & 0.8187                                                    & 0.8493 & 0.8536 & 0.8445 \\
@edge  & 0.6405                                                    & 0.6894 & 0.6993 & 0.6731 \\
@in    & 0.8407                                                    & 0.8692 & 0.8729 & 0.8658 \\ \bottomrule
\end{tabular} }
\end{table}

\begin{figure}[t]
  \includegraphics[width=0.45\textwidth]{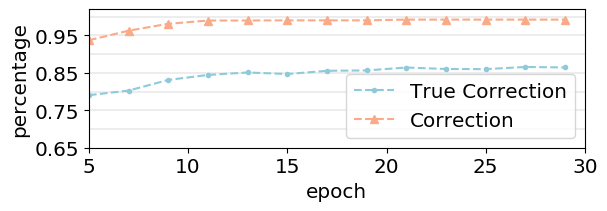}
  \caption{The percentages of points
  with replaced label (denoted as Correction) and the percentage of correctly corrected points (denoted as True Correction). The correction gradually explores the entire dataset.}
  \label{Fig:corrector_percent}
  \end{figure}
  
\begin{figure*}[h] 
  \centering
  \setlength{\tabcolsep}{0.15mm}{
  \begin{tabular}{cccccc}
  \multicolumn{6}{c}{\includegraphics[width=1.0\textwidth]{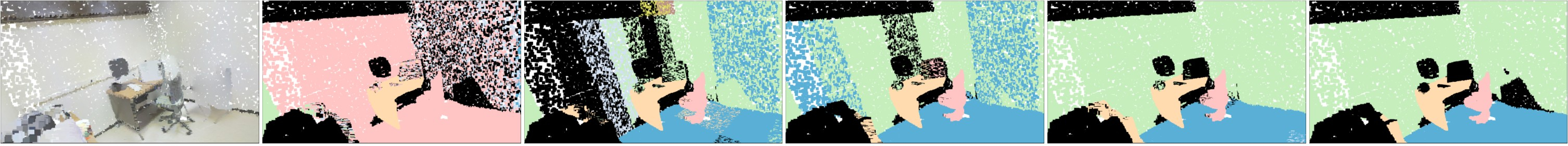}} \\ 
  
  \multicolumn{6}{c}{\includegraphics[width=1.0\textwidth]{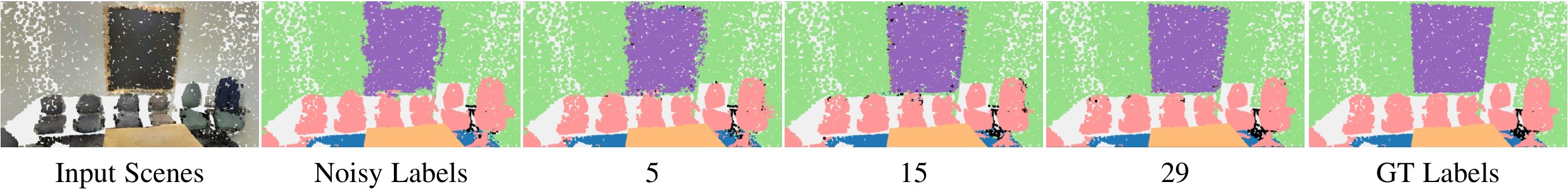}}
  \\ 
  
  \end{tabular}
  }
  \caption{Comparison of the corrected labels in epochs 5, 15, and 29 on the $\alpha=1.0, \beta=0.7$ noisy S3DIS. From left to right are the input point cloud with RGB, the corresponding noisy label, the label correction in epochs 5, 15, and 29, and the ground-truth label. \ysq{The first row is on instance noise S3DIS and the second row is on boundary noise S3DIS.}}
  \label{Fig:s3dis_corrprocess}
  \end{figure*}

\noindent\textbf{Robustness to Clustering Methods and Granularity.}\label{subsec:abl2_vsEpsandGMM}
In Table~\ref{tab:granularity}, the first three columns report the results of PNAL under different clustering granularities ($\varepsilon$ is $0.015, 0.018, 0.021$), and the last two columns report the results of ours under other type of clustering (GMM, spectral). They show close results, which demonstrates that our method is robust to a certain range of cluster granularities and is not sensitive to the clustering method used.

\frysq{\noindent\textbf{Iterative Cleaning.}
In Table~\ref{tab:iterative}, we report the results of iteratively applying PNAL within the total 30 training epochs, where the prediction from the previous iteration is used to train new models and perform cleaning in the next iteration. Each column shows different schedules, i.e., at which epochs we start a new iteration. The iteration start at epochs 15 gives a similar performance compared to ours default setting, while the ones iterated more frequently show a performance decline, which may be due to inadequate learning and cleaning.
}

\subsubsection{Ablation on PNAL-boundary}

  \begin{figure*}[t] 
  \centering
  \setlength{\tabcolsep}{0.2mm}{
  \begin{tabular}{ccccccc}
  \includegraphics[width=0.142\textwidth,page=1]{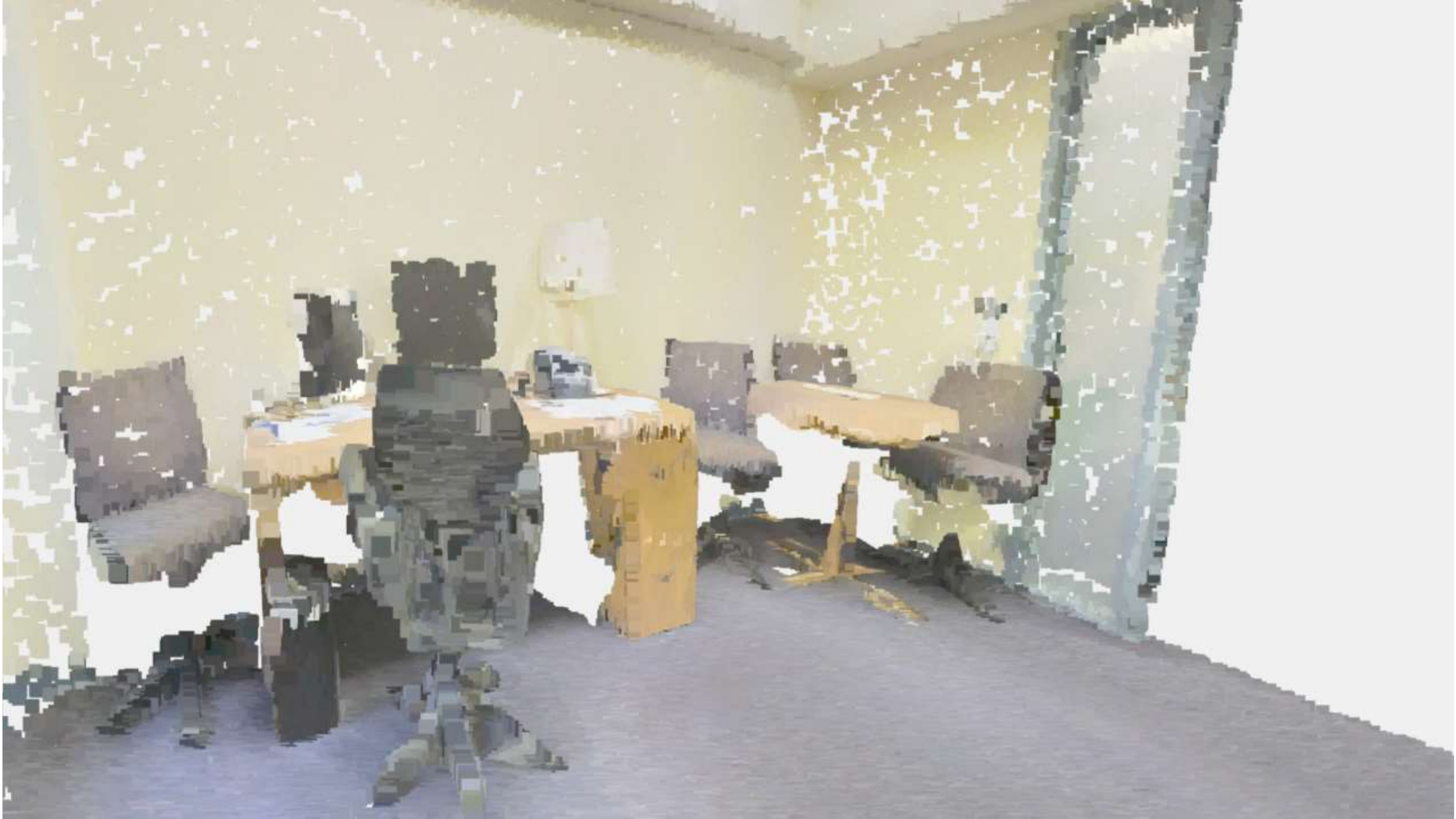} &
  \includegraphics[width=0.142\textwidth,page=2]{LaTeX/img/corrprocess/drawprocess_mixnoise.pdf} &
  \includegraphics[width=0.142\textwidth,page=3]{LaTeX/img/corrprocess/drawprocess_mixnoise.pdf} & 
  \includegraphics[width=0.142\textwidth,page=4]{LaTeX/img/corrprocess/drawprocess_mixnoise.pdf} & 
  \includegraphics[width=0.142\textwidth,page=5]{LaTeX/img/corrprocess/drawprocess_mixnoise.pdf} &
  \includegraphics[width=0.142\textwidth,page=6]{LaTeX/img/corrprocess/drawprocess_mixnoise.pdf} &
  \includegraphics[width=0.142\textwidth,page=6]{LaTeX/img/corrprocess/drawprocess_mixnoise.pdf}
  \\ 
    Input Scenes & Noisy Labels & 5 & 15 & 29 & 39 & GT Labels\\
  
  \end{tabular}
  }
  \caption{Comparison of the corrected labels in epochs 5, 15, 29, and 39 on the $\alpha=1.0, \beta=0.7$ mixed noise S3DIS. 5, 15, 29 are the instance-level corrected labels by PNAL, and 39 is the boundary-level corrected labels by PNAL-boundary.}
  \label{Fig:s3dis_corrprocess_mix}
  \end{figure*}

\noindent\textbf{Component Ablation Study.}
\ysq{All the results in Table. \ref{tab:ablation_boundary} are on $\alpha=1.0, \beta=0.7$ boundary noise.
The first column reports the results of PNAL-boundary without the progressive boundary correction. Instead of obtaining new boundary and neighbor points from the latest cleaned ground truth, here we obtain the boundary and neighbor points from the initial ground truth label. 
Compared with our progressive way, the non-progressive way has a small drop, which can illustrate the feasibility of progressive boundary correction based on the latest cleaned ground truth. }

\noindent\textbf{Robustness to $k$.}
\ysq{In this study, we show the robustness of PNAL-boundary on the number of nearest neighbors of the progressive boundary correction, which is denoted as $k$. This parameter controls the range of boundary points whose ground-truth label can be corrected within each epoch.
In the second, third and forth columns in Table. \ref{tab:ablation_boundary}, we adjust $k$ to 10 and 30. We use $k=20$ in our default setting. Results show a little performance drop with a different $k$, implying our method is not sensitive to the choice of $k$. Also, we find that the performance drops when $k$ increases. It may be due to the incorrect correction of the non-boundary point, when the range of boundary points is too large.}

\subsection{Analysis of Correction Process }\label{subsec:corrprocess}

\subsubsection{Instance-level Label Noise}
The first row of Fig. \ref{Fig:s3dis_corrprocess} shows the visualization of the label correction process by the PNAL during training. We can find that as the training goes, the overall labeling errors in the training set tends to decrease and gradually approaches the clean ground-truth label. Typically, label errors with large areas (e.g., floors, walls, ceilings) are corrected first. As training proceeds, PNAL gradually explores the entire dataset and tries to correct difficult and small objects. As given in Fig. \ref{Fig:corrector_percent}, the percentage of points with replaced label increases from $0.936$ to $0.992$, while the percentage of correctly corrected points is close to $0.8$ from the beginning of the noise-cleaning stage and then gradually increases to $0.865$. It also shows that the PNAL correction process spreads to the whole training set as the training proceeds. To note that we take the case of correcting to the original label into account.

\subsubsection{Boundary-level Label Noise} 

\ysq{The second row of Fig. \ref{Fig:s3dis_corrprocess} shows the visualization of the label correction process by the PNAL-boundary during training.
We can find that, as the training goes, the proposed PNAL-boundary framework corrects the coarsely annotated boundary labels to align with the real boundary between the wall and the board in a coarse-to-fine manner. For most of the inner points, their labels are protected from false corrections.}

\subsubsection{Mixed Label Noise} \label{subsec:corrprocess_mixed}

\ysq{Fig. \ref{Fig:s3dis_corrprocess_mix} shows the visualization of the label correction process on mixed boundary and instance label noise by the our frameworks during training. The first 30 epoches are trained with PNAL framework and the last 10 epoches are with PNAL-boundary.
In the first 30 epochs, as the training goes, the PNAL framework corrects the large areas first (e.g., wall), and then small objects (e.g., big table, small table, chair), which is consistent with our previous observations. However, false correction appears at not only boundary, but also inner points (e.g., the smaller item on the table is labeled as table at 29 epoch).
In the last 10 epoches, the proposed PNAL-boundary framework corrects the coarsely annotated boundary labels to align with the real boundary between the wall and the board in a coarse-to-fine manner. While the labels of the inner points are protected from false corrections, however, we can find that the true corrections are also limited to a certain range. For example, the smaller item (in black color) on the table, which has been incorrectly relabeled as table (in desert color) by PNAL, cannot be fully corrected by PNAL-boundary in a limited steps.}

\section{Conclusion}
\lj{In this study, we propose a novel point cloud segmentation framework PNAL and its variant PNAL-boundary, to cope with the point cloud labeling noise problem at both instance and boundary levels.} Unlike existing methods that focus on image classification, PNAL and PNAL-boundary are noise-rate blind, in order to deal with the unique noise-rate variation problem in point cloud. We propose point-wise confidence selection, cluster-wise label correction and voting strategies to generate the best possible labels considering the correlation labels in local similar points ysq{for instance-level label noise. For boundary label noise, we propose a progressive boundary label cleaning strategy.} In addition, we re-labeled the validation set of a popular but noisy real-world scene point cloud dataset to make it clean, for rigorous experiment and for future research. Experiments demonstrate the effectiveness and robustness of our method on real-world noisy data and artificially created noisy public data. \frysq{For future directions, as the boundary detection models~\cite{gong2021boundary} trained on boundary-noisy labels tend to produce thicker boundaries, as reported in~\cite{AcunaCVPR19STEAL}, it would be a possible to utilize boundary detection models to select possible noisy labeled boundaries from thick predictions for further cleaning.}

\ifCLASSOPTIONcompsoc
  \section*{Acknowledgments}
\else
  \section*{Acknowledgment}
\fi

We would like to thank Jiaying Lin for the discussions and for building the annotation site for the proposed dataset. This work was supported by the HKSAR Innovation and Technology Commission (ITC) under ITF Project MHP/109/19.

\ifCLASSOPTIONcaptionsoff
  \newpage
\fi

\bibliographystyle{IEEEtran}
\bibliography{bib}

\begin{thebibliography}{10}
\providecommand{\url}[1]{#1}
\csname url@samestyle\endcsname
\providecommand{\newblock}{\relax}
\providecommand{\bibinfo}[2]{#2}
\providecommand{\BIBentrySTDinterwordspacing}{\spaceskip=0pt\relax}
\providecommand{\BIBentryALTinterwordstretchfactor}{4}
\providecommand{\BIBentryALTinterwordspacing}{\spaceskip=\fontdimen2\font plus
\BIBentryALTinterwordstretchfactor\fontdimen3\font minus
  \fontdimen4\font\relax}
\providecommand{\BIBforeignlanguage}[2]{{%
\expandafter\ifx\csname l@#1\endcsname\relax
\typeout{** WARNING: IEEEtran.bst: No hyphenation pattern has been}%
\typeout{** loaded for the language `#1'. Using the pattern for}%
\typeout{** the default language instead.}%
\else
\language=\csname l@#1\endcsname
\fi
#2}}
\providecommand{\BIBdecl}{\relax}
\BIBdecl

\bibitem{Cordts2016Cityscapes}
M.~Cordts, M.~Omran, S.~Ramos, T.~Rehfeld, M.~Enzweiler, R.~Benenson,
  U.~Franke, S.~Roth, and B.~Schiele, ``The cityscapes dataset for semantic
  urban scene understanding,'' in \emph{Proc. of the IEEE Conference on
  Computer Vision and Pattern Recognition (CVPR)}, 2016.

\bibitem{Tan_2021_TIP_NightCity}
X.~Tan, K.~Xu, Y.~Cao, Y.~Zhang, L.~Ma, and R.~W.~H. Lau, ``Night-time scene
  parsing with a large real dataset,'' \emph{IEEE Transactions on Image
  Processing}, vol.~30, pp. 9085--9098, 2021.

\bibitem{dai2017scannet}
A.~Dai, A.~X. Chang, M.~Savva, M.~Halber, T.~Funkhouser, and M.~Nie{\ss}ner,
  ``Scannet: Richly-annotated 3d reconstructions of indoor scenes,'' in
  \emph{Proc. Computer Vision and Pattern Recognition (CVPR), IEEE}, 2017.

\bibitem{10.5555/3327757.3327944}
B.~Han, Q.~Yao, X.~Yu, G.~Niu, M.~Xu, W.~Hu, I.~W. Tsang, and M.~Sugiyama,
  ``Co-teaching: Robust training of deep neural networks with extremely noisy
  labels,'' in \emph{Proceedings of the 32nd International Conference on Neural
  Information Processing Systems}, ser. NIPS'18.\hskip 1em plus 0.5em minus
  0.4em\relax Red Hook, NY, USA: Curran Associates Inc., 2018, p. 8536–8546.

\bibitem{yu2019does}
X.~Yu, B.~Han, J.~Yao, G.~Niu, I.~Tsang, and M.~Sugiyama, ``How does
  disagreement help generalization against label corruption?'' in
  \emph{International Conference on Machine Learning}, 2019, pp. 7164--7173.

\bibitem{shen2019learning}
Y.~Shen and S.~Sanghavi, ``Learning with bad training data via iterative
  trimmed loss minimization,'' in \emph{International Conference on Machine
  Learning}.\hskip 1em plus 0.5em minus 0.4em\relax PMLR, 2019, pp. 5739--5748.

\bibitem{song2019selfie}
H.~Song, M.~Kim, and J.-G. Lee, ``Selfie: Refurbishing unclean samples for
  robust deep learning,'' in \emph{International Conference on Machine
  Learning}.\hskip 1em plus 0.5em minus 0.4em\relax PMLR, 2019, pp. 5907--5915.

\bibitem{liu2020early}
S.~Liu, J.~Niles-Weed, N.~Razavian, and C.~Fernandez-Granda, ``Early-learning
  regularization prevents memorization of noisy labels,'' \emph{Advances in
  Neural Information Processing Systems}, vol.~33, 2020.

\bibitem{zhang2018generalized}
Z.~Zhang and M.~R. Sabuncu, ``Generalized cross entropy loss for training deep
  neural networks with noisy labels,'' in \emph{NeurIPS}, 2018.

\bibitem{Wang_2019_ICCV}
Y.~Wang, X.~Ma, Z.~Chen, Y.~Luo, J.~Yi, and J.~Bailey, ``Symmetric cross
  entropy for robust learning with noisy labels,'' in \emph{Proceedings of the
  IEEE/CVF International Conference on Computer Vision (ICCV)}, October 2019.

\bibitem{Reed2015TrainingDN}
S.~Reed, H.~Lee, D.~Anguelov, C.~Szegedy, D.~Erhan, and A.~Rabinovich,
  ``Training deep neural networks on noisy labels with bootstrapping,''
  \emph{ICLR}, vol. abs/1412.6596, 2015.

\bibitem{ICML2019_UnsupervisedLabelNoise}
E.~Arazo, D.~Ortego, P.~Albert, N.~E. O'Connor, and K.~McGuinness,
  ``Unsupervised label noise modeling and loss correction,'' in
  \emph{International Conference on Machine Learning (ICML)}, June 2019.

\bibitem{armeni_cvpr16}
I.~Armeni, O.~Sener, A.~R. Zamir, H.~Jiang, I.~Brilakis, M.~Fischer, and
  S.~Savarese, ``3d semantic parsing of large-scale indoor spaces,'' in
  \emph{Proceedings of the IEEE International Conference on Computer Vision and
  Pattern Recognition}, 2016.

\bibitem{qi2017pointnet}
C.~R. Qi, H.~Su, K.~Mo, and L.~J. Guibas, ``Pointnet: Deep learning on point
  sets for 3d classification and segmentation,'' in \emph{Proceedings of the
  IEEE conference on computer vision and pattern recognition}, 2017, pp.
  652--660.

\bibitem{Ye2021MetaPUAA}
S.~Ye, D.~Chen, S.~Han, Z.~Wan, and J.~Liao, ``Meta-pu: An arbitrary-scale
  upsampling network for point cloud,'' \emph{IEEE transactions on
  visualization and computer graphics}, vol.~PP, 2021.

\bibitem{ye20213d}
S.~Ye, D.~Chen, S.~Han, and J.~Liao, ``3d question answering,'' 2021.

\bibitem{han2021exemplarbased}
F.~Han, S.~Ye, M.~He, M.~Chai, and J.~Liao, ``Exemplar-based 3d portrait
  stylization,'' \emph{IEEE Transactions on Visualization and Computer
  Graphics}, 2021.

\bibitem{gong2021boundary}
J.~Gong, J.~Xu, X.~Tan, J.~Zhou, Y.~Qu, Y.~Xie, and L.~Ma, ``Boundary-aware
  geometric encoding for semantic segmentation of point clouds,'' in
  \emph{Proceedings of the AAAI Conference on Artificial Intelligence},
  vol.~35, no.~2, 2021, pp. 1424--1432.

\bibitem{tatarchenko2017octree}
M.~Tatarchenko, A.~Dosovitskiy, and T.~Brox, ``Octree generating networks:
  Efficient convolutional architectures for high-resolution 3d outputs,'' in
  \emph{Proceedings of the IEEE International Conference on Computer Vision},
  2017, pp. 2088--2096.

\bibitem{klokov2017escape}
R.~Klokov and V.~Lempitsky, ``Escape from cells: Deep kd-networks for the
  recognition of 3d point cloud models,'' in \emph{Proceedings of the IEEE
  International Conference on Computer Vision}, 2017, pp. 863--872.

\bibitem{landrieu2018large}
L.~Landrieu and M.~Simonovsky, ``Large-scale point cloud semantic segmentation
  with superpoint graphs,'' in \emph{Proceedings of the IEEE Conference on
  Computer Vision and Pattern Recognition}, 2018, pp. 4558--4567.

\bibitem{7298885}
{Tong Xiao}, {Tian Xia}, {Yi Yang}, {Chang Huang}, and {Xiaogang Wang},
  ``Learning from massive noisy labeled data for image classification,'' in
  \emph{2015 IEEE Conference on Computer Vision and Pattern Recognition
  (CVPR)}, 2015, pp. 2691--2699.

\bibitem{Goldberger2017TrainingDN}
J.~Goldberger and E.~Ben-Reuven, ``Training deep neural-networks using a noise
  adaptation layer,'' in \emph{ICLR}, 2017.

\bibitem{7472164}
A.~J. {Bekker} and J.~{Goldberger}, ``Training deep neural-networks based on
  unreliable labels,'' in \emph{2016 IEEE International Conference on
  Acoustics, Speech and Signal Processing (ICASSP)}, 2016, pp. 2682--2686.

\bibitem{ghosh2017robust}
A.~Ghosh, H.~Kumar, and P.~Sastry, ``Robust loss functions under label noise
  for deep neural networks,'' in \emph{Proceedings of the AAAI Conference on
  Artificial Intelligence}, vol.~31, no.~1, 2017.

\bibitem{DBLP:conf/iclr/LyuT20}
\BIBentryALTinterwordspacing
Y.~Lyu and I.~W. Tsang, ``Curriculum loss: Robust learning and generalization
  against label corruption,'' in \emph{8th International Conference on Learning
  Representations, {ICLR} 2020, Addis Ababa, Ethiopia, April 26-30,
  2020}.\hskip 1em plus 0.5em minus 0.4em\relax OpenReview.net, 2020. [Online].
  Available: \url{https://openreview.net/forum?id=rkgt0REKwS}
\BIBentrySTDinterwordspacing

\bibitem{yang2020lncis}
Y.~Longrong, M.~Fanman, L.~Hongliang, W.~Qingbo, and C.~Qishang, ``Learning
  with noisy class labels for instance segmentation,'' in \emph{European
  Conference on Computer Vision (ECCV)}, 2020.

\bibitem{NIPS2017_2f37d101}
\BIBentryALTinterwordspacing
H.-S. Chang, E.~Learned-Miller, and A.~McCallum, ``Active bias: Training more
  accurate neural networks by emphasizing high variance samples,'' in
  \emph{Advances in Neural Information Processing Systems}, I.~Guyon, U.~V.
  Luxburg, S.~Bengio, H.~Wallach, R.~Fergus, S.~Vishwanathan, and R.~Garnett,
  Eds., vol.~30.\hskip 1em plus 0.5em minus 0.4em\relax Curran Associates,
  Inc., 2017. [Online]. Available:
  \url{https://proceedings.neurips.cc/paper/2017/file/2f37d10131f2a483a8dd005b3d14b0d9-Paper.pdf}
\BIBentrySTDinterwordspacing

\bibitem{NEURIPS2018_ad554d8c}
\BIBentryALTinterwordspacing
D.~Hendrycks, M.~Mazeika, D.~Wilson, and K.~Gimpel, ``Using trusted data to
  train deep networks on labels corrupted by severe noise,'' in \emph{Advances
  in Neural Information Processing Systems}, S.~Bengio, H.~Wallach,
  H.~Larochelle, K.~Grauman, N.~Cesa-Bianchi, and R.~Garnett, Eds.,
  vol.~31.\hskip 1em plus 0.5em minus 0.4em\relax Curran Associates, Inc.,
  2018. [Online]. Available:
  \url{https://proceedings.neurips.cc/paper/2018/file/ad554d8c3b06d6b97ee76a2448bd7913-Paper.pdf}
\BIBentrySTDinterwordspacing

\bibitem{zhang2020characterizing}
M.~Zhang, J.~Gao, Z.~Lyu, W.~Zhao, Q.~Wang, W.~Ding, S.~Wang, Z.~Li, and
  S.~Cui, ``Characterizing label errors: Confident learning for noisy-labeled
  image segmentation,'' in \emph{International Conference on Medical Image
  Computing and Computer-Assisted Intervention}.\hskip 1em plus 0.5em minus
  0.4em\relax Springer, 2020, pp. 721--730.

\bibitem{wang2020noise}
G.~Wang, X.~Liu, C.~Li, Z.~Xu, J.~Ruan, H.~Zhu, T.~Meng, K.~Li, N.~Huang, and
  S.~Zhang, ``A noise-robust framework for automatic segmentation of covid-19
  pneumonia lesions from ct images,'' \emph{IEEE Transactions on Medical
  Imaging}, vol.~39, no.~8, pp. 2653--2663, 2020.

\bibitem{pnal2021}
S.~Ye, D.~Chen, S.~Han, and J.~Liao, ``Learning with noisy labels for robust
  point cloud segmentation,'' \emph{International Conference on Computer
  Vision}, 2021.

\bibitem{yu2018seal}
Z.~Yu, W.~Liu, Y.~Zou, C.~Feng, S.~Ramalingam, B.~V. K.~V. Kumar, and J.~Kautz,
  ``Simultaneous edge alignment and learning,'' in \emph{European Conference on
  Computer Vision (ECCV)}, 2018.

\bibitem{AcunaCVPR19STEAL}
D.~Acuna, A.~Kar, and S.~Fidler, ``Devil is in the edges: Learning semantic
  boundaries from noisy annotations,'' in \emph{CVPR}, 2019.

\bibitem{zhang2018deep}
J.~Zhang, T.~Zhang, Y.~Dai, M.~Harandi, and R.~Hartley, ``Deep unsupervised
  saliency detection: A multiple noisy labeling perspective,'' in \emph{Proc.
  CVPR}, 2018, pp. 9029--9038.

\bibitem{PMD:2020}
J.~Lin, G.~Wang, and R.~W. Lau, ``Progressive mirror detection,'' in
  \emph{Proc. CVPR}, 2020.

\bibitem{GSD:2021}
J.~Lin, Z.~He, and R.~W. Lau, ``Rich context aggregation with reflection prior
  for glass surface detection,'' in \emph{Proc. CVPR}, 2021.

\bibitem{zhu2019pick}
H.~Zhu, J.~Shi, and J.~Wu, ``Pick-and-learn: automatic quality evaluation for
  noisy-labeled image segmentation,'' in \emph{International Conference on
  Medical Image Computing and Computer-Assisted Intervention}.\hskip 1em plus
  0.5em minus 0.4em\relax Springer, 2019, pp. 576--584.

\bibitem{xue2020cas}
C.~Xue, Q.~Deng, X.~Li, Q.~Dou, and P.-A. Heng, ``Cascaded robust learning at
  imperfect labels for chest x-ray segmentation,'' in \emph{International
  conference on medical image computing and computer-assisted
  intervention}.\hskip 1em plus 0.5em minus 0.4em\relax Springer, 2020, pp.
  579--588.

\bibitem{zhang2020robust}
T.~Zhang, L.~Yu, N.~Hu, S.~Lv, and S.~Gu, ``Robust medical image segmentation
  from non-expert annotations with tri-network,'' in \emph{International
  Conference on medical image computing and computer-assisted
  intervention}.\hskip 1em plus 0.5em minus 0.4em\relax Springer, 2020, pp.
  249--258.

\bibitem{shi2021distilling}
J.~Shi and J.~Wu, ``Distilling effective supervision for robust medical image
  segmentation with noisy labels,'' in \emph{International Conference on
  Medical Image Computing and Computer-Assisted Intervention}.\hskip 1em plus
  0.5em minus 0.4em\relax Springer, 2021, pp. 668--677.

\bibitem{li2021superpixel}
S.~Li, Z.~Gao, and X.~He, ``Superpixel-guided iterative learning from noisy
  labels for medical image segmentation,'' in \emph{International Conference on
  Medical Image Computing and Computer-Assisted Intervention}.\hskip 1em plus
  0.5em minus 0.4em\relax Springer, 2021, pp. 525--535.

\bibitem{ester1996density}
M.~Ester, H.-P. Kriegel, J.~Sander, and X.~Xu, ``Density-based spatial
  clustering of applications with noise,'' in \emph{Int. Conf. Knowledge
  Discovery and Data Mining}, vol. 240, 1996, p.~6.

\bibitem{arpit2017closer}
D.~Arpit, S.~Jastrz{\k{e}}bski, N.~Ballas, D.~Krueger, E.~Bengio, M.~S. Kanwal,
  T.~Maharaj, A.~Fischer, A.~Courville, Y.~Bengio \emph{et~al.}, ``A closer
  look at memorization in deep networks,'' in \emph{International Conference on
  Machine Learning}.\hskip 1em plus 0.5em minus 0.4em\relax PMLR, 2017, pp.
  233--242.

\bibitem{smartscenes}
\BIBentryALTinterwordspacing
Smartscenes, ``smartscenes/sstk.'' [Online]. Available:
  \url{https://github.com/smartscenes/sstk/wiki/Scan-Annotation-Pipeline}
\BIBentrySTDinterwordspacing

\bibitem{felzenszwalb2004efficient}
P.~F. Felzenszwalb and D.~P. Huttenlocher, ``Efficient graph-based image
  segmentation,'' \emph{International journal of computer vision}, vol.~59,
  no.~2, pp. 167--181, 2004.

\bibitem{pham-jsis3d-cvpr19}
Q.-H. Pham, D.~T. Nguyen, B.-S. Hua, G.~Roig, and S.-K. Yeung, ``{JSIS3D}:
  Joint semantic-instance segmentation of 3d point clouds with multi-task
  pointwise networks and multi-value conditional random fields,'' in
  \emph{Proceedings of the IEEE Conference on Computer Vision and Pattern
  Recognition (CVPR)}, 2019.

\bibitem{dgcnn}
Y.~Wang, Y.~Sun, Z.~Liu, S.~E. Sarma, M.~M. Bronstein, and J.~M. Solomon,
  ``Dynamic graph cnn for learning on point clouds,'' \emph{ACM Transactions on
  Graphics (TOG)}, 2019.

\bibitem{qi2017pointnetplusplus}
C.~R. Qi, L.~Yi, H.~Su, and L.~J. Guibas, ``Pointnet++: Deep hierarchical
  feature learning on point sets in a metric space,'' \emph{arXiv preprint
  arXiv:1706.02413}, 2017.

\bibitem{3DSemanticSegmentationWithSubmanifoldSparseConvNet}
B.~Graham, M.~Engelcke, and L.~van~der Maaten, ``3d semantic segmentation with
  submanifold sparse convolutional networks,'' \emph{CVPR}, 2018.

\end{thebibliography}

\end{document}